%% file: 0_main.tex
\PassOptionsToPackage{warn}{textcomp}
\documentclass[sigconf]{acmart}
\usepackage{booktabs} 
\usepackage{times}
\usepackage{epsfig}
\usepackage{graphicx}
\usepackage{amsmath}
\usepackage{amssymb}
\usepackage{amsmath,}
\usepackage{algorithm}
\usepackage[noend]{algpseudocode}
\usepackage{pseudocode}
\usepackage{multirow}
\usepackage{caption}
\usepackage{hhline}
\usepackage{xcolor}
\usepackage{dblfloatfix}
\usepackage{subcaption}
\usepackage{enumitem}
\usepackage{ulem}
\usepackage{color, colortbl}
\usepackage{sidecap}
\sidecaptionvpos{figure}{c}

\setcopyright{none}

\acmDOI{10.1145/3377930.3390226}
\acmISBN{978-1-4503-7128-5/20/07}

\acmConference[GECCO '20]{the Genetic and Evolutionary Computation Conference 2020}{July 8--12, 2020}{Cancun, Mexico}
\acmYear{2020}

\definecolor{lightblue}{rgb}{0.8,0.93,1}
\newcommand{\emptycell}{\cellcolor{white}}
\newcommand{\hlblue}[1]{\colorbox{lightblue}{#1}}
\newcommand{\sys}{GeneCAI} 
\newcommand{\mli}[1]{\mathit{#1}}

\newcommand{\eq}[1]{Eq.~({#1})}

\graphicspath{{figures/}} 

\newcolumntype{b}{>{\columncolor{lightblue}}c}

\begin{document}

\setlength{\belowdisplayskip}{3pt} \setlength{\belowdisplayshortskip}{3pt}
\setlength{\abovedisplayskip}{3pt} \setlength{\abovedisplayshortskip}{3pt}
\setlength{\textfloatsep}{1pt}

\title{\sys{}: \underline{Gene}tic Evolution for Acquiring \underline{C}ompact \underline{AI}
}

\author{Mojan Javaheripi}
\affiliation{%
  \institution{ECE Dept., UCSD, USA}
}
\email{mojan@ucsd.edu}
\author{Mohammad Samragh}
\affiliation{%
  \institution{ECE Dept., UCSD, USA}
}
\email{msamragh@ucsd.edu}
\author{Tara Javidi}
\affiliation{%
  \institution{ECE Dept., UCSD, USA}
}
\email{tjavidi@ucsd.edu}
\author{Farinaz Koushanfar}
\affiliation{%
  \institution{ECE Dept., UCSD, USA}
}
\email{farinaz@ucsd.edu}


\begin{abstract}
In the contemporary big data realm, Deep Neural Networks (DNNs) are evolving towards more complex architectures to achieve higher inference accuracy. Model compression techniques can be leveraged to efficiently deploy such compute-intensive architectures on resource-limited mobile devices. Such methods comprise various hyperparameters that require per-layer customization to ensure high accuracy. Choosing such hyperparameters is cumbersome as the pertinent search space grows exponentially with model layers. This paper introduces \sys{}, a novel optimization method that automatically \textit{learn}s how to tune per-layer compression hyperparameters. We devise a bijective translation scheme that encodes compressed DNNs to the \textit{genotype} space. Each genotype's optimality is measured using a multi-objective score based on accuracy and number of floating point operations. We develop customized genetic operations to iteratively evolve the non-dominated solutions towards the optimal Pareto front, thus, capturing the optimal trade-off between model accuracy and complexity. \sys{} optimization method is highly scalable and can achieve a near-linear performance boost on distributed multi-GPU platforms. Our extensive evaluations demonstrate that \sys{} outperforms existing rule-based and reinforcement learning methods in DNN compression by finding models that lie on a better accuracy/complexity Pareto curve.
\end{abstract}

%
%
\vspace{-1cm}


\keywords{
\vspace{-0.1cm}Deep Learning, Genetic Algorithms, Multi-objective Optimization, Computer aided/automated design, Parallel Optimization}

\maketitle

\vspace{-0.3cm}
\input{1_intro}
\input{2_related}
\input{3_formulation}
\input{4_approach}
\input{5_experiments}
\input{6_conclusion}

\bibliographystyle{ACM-Reference-Format}
\bibliography{sample-bibliography}

\end{document}

%% file: 1_intro.tex
\section{Introduction}\label{sec:intro}
With the growing range of applications for Deep Neural Networks (DNNs), the demand for higher accuracy has led to a continuous increase in the complexity of state-of-the-art models. Such high execution cost hinders the deployment of DNNs in real-time applications on commodity hardware. Fortunately, modern neural networks have been shown to incur high redundancies that can be eliminated without compromising inference accuracy~\cite{javaheripi2019swnet}. 
Effective identification and removal of such redundancies has fueled research in two interlinked domains: (i)~Developing model compression techniques, e.g., pruning~\cite{lin2018accelerating,jiang2018efficient,he2018soft,he2017channel,li2016pruning}, quantization~\cite{zhou2016dorefa}, and coding~\cite{han2015deep}. (ii)~Devising automated policies that learn how to configure compression techniques to jointly achieve accuracy and compactness~\cite{wang2018haq,he2018amc,samragh2019codex,yazdanbakhsh2018releq, samragh2019autorank,javaheripi2019peeking}. In this paper, we focus on the latter.

\begin{SCfigure}[50][ht]
\centering
\includegraphics[width=0.5\columnwidth]{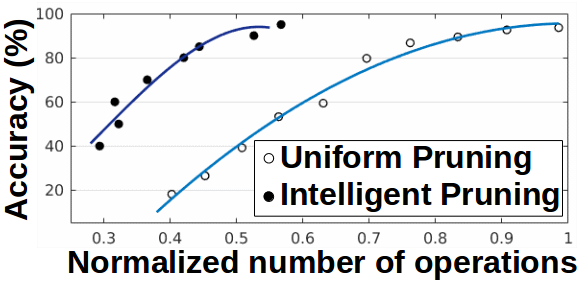}
\caption{Accuracy/FLOPs Pareto curves for pruning a pre-trained VGG network on CIFAR-10 benchmark. 
    }\label{fig:pareto}
\vspace{-0.2cm}
\end{SCfigure}

The effectiveness of contemporary compression techniques relies on careful tuning of several hyperparameters across DNN layers. These hyperparameters directly control the trade-off between accuracy and execution cost on a constrained device. The question to be answered is how to find an optimal hyperparameter configuration that results in a high compression rate while minimally affecting inference accuracy. Figure~\ref{fig:pareto} shows how an intelligent hyperparameter selection policy can better estimate the geometry of the optimal Pareto front for the same compression technique (Pruning).

\vspace{-0.2cm}

Finding the optimal set of hyperparameters is quite challenging as the space of possibilities grows exponentially with number of DNN layers. 
Such large search-space renders manual or computerized greedy algorithms sub-optimal or infeasible for hyperparameter tuning. 
Existing work in automated compression suggests the use of heuristic methods~\cite{he2018soft,he2017channel,jiang2018efficient} or Reinforcement Learning (RL)~\cite{he2018amc,yazdanbakhsh2018releq,yang2018netadapt}. To tackle the high-dimensionality of the search space, heuristics and RL algorithms specify the hyperparameters one layer at a time, thus overlooking the layer inter-dependencies.

In this paper, we take an alternative approach, called \sys{}, that \textit{simultaneously} tunes the compression hyperparameters for all DNN layers using genetic algorithms (GA). We devise a customized translator that encapsulates the hyperparameter configuration of all DNN layers as \textit{one} fixed-length vector $\vec x\in\mathbb{R}^d$, called an \textit{individual}. Using this translation scheme, each hyperparameter vector translates to a unique compressed DNN.
In this setting, the search for compression hyperparameters directly translates to optimizing an objective function $f(\vec x):\mathbb{R}^d\rightarrow{\mathbb{R}}$ over $\vec x$ where $f(\cdot)$ is an arbitrary measure of quality. 
We suggest a unified formulation for $f(\cdot)$, called the \textit{score}, which assesses the quality of individuals by combining inference accuracy and a desired execution cost, e.g., runtime.

\sys{} iteratively \textit{evolve}s the initial \textit{population} towards superior individuals. To promote high-quality initialization, we propose a context-aware boundary characterization that significantly contributes to convergence and search results. After initialization, each evolution iteration comprises three consecutive genetic operations, namely, \textit{selection}, \textit{crossover}, and \textit{mutation}. We balance \sys{} exploration and exploitation by devising a diversity-guided GA. In particular, we define and measure diversity at the genotype level and adapt our crossover and mutation operations accordingly. By maintaining a diversified population, \sys{} ensures a fast and stable search and prevents premature convergence.

In high-level, \sys{} offers several benefits: \textbf{(I)}~it is highly cost-effective as it takes a post-processing approach where a pre-trained DNN is compressed while no training is required amid GA iterations. \textbf{(II)}~\sys{} exploits parallelism to reduce optimization time by concurrent sample evaluations; We develop an accompanying API that manages the infrastructure for multi-GPU execution of \sys{} search. \textbf{(III)}~\sys{} makes no assumptions about the objective function; by adjusting the cost, one can apply \sys{} to various DNN compression tasks and hardware constraints. \textbf{(IV)}~\sys{} encoding and genetic operations accommodate optimization of both discrete and continuous-valued hyperparameters.

To demonstrate \sys{} generalizability, we evaluate our methodology on four compression techniques and several architectures.
Our evaluations unveil the full potential of GA in learning to effectively combine multiple methods and push the limits of network compression. For VGG-16 on ImageNet, \sys{} pushes the state-of-the-art FLOPs reduction from $5\times$ to $7.1\times$ with higher accuracy.
For expert-designed efficient MobileNets, \sys{} obtains on average $1.2\%$ higher top-1 accuracy than the MobileNet Pareto curve~\cite{howard2017mobilenets}. We further show that \sys{} computational flow is highly scalable and enjoys a linear reduction in search time with an increased number of distributed computing resources.

%% file: 2_related.tex
\section{Background and Related Work}\label{sec:related}
\noindent\textbf{Neural Network Compression. }Existing model compression techniques can be roughly categorized in two classes: post-training compression and compression-aware training. The post-training approach is applied to pre-trained DNNs, without modifying the standard training loss or the back-propagation routines~\cite{he2018soft,he2017channel,wang2017structured,jiang2018efficient,li2016pruning,lin2017runtime,lin2018accelerating,luo2017thinet,kim2015compression,samragh2017customizing}.
Alternatively, in compression-aware training~\cite{alvarez2017compression,ghasemzadeh2018rebnet}, the training loss or back-propagation algorithm is altered to create models that are inherently \textit{compressible}.

Each compression method has its own benefits and tradeoffs; In either case, the compression hyperparameters still need to be specified. 
Most of the existing work in DNN compression literature rely on hand-crafted or heuristic methods for per-layer configuration of the compression hyperparameters, which can be cumbersome and result in sub-optimal solutions. Different from heuristic methods that compress one layer at a time, \sys{} performs whole-network compression where all layers are simultaneously optimized to account for inter-layer correlations. We target post-training compression and apply fine-tuning only once. This allows us to customize DNNs for various hardware platforms with little search/training overhead. 

\noindent\textbf{Automated Architecture Search. } Modifying the DNN architecture graph in search for higher training/inference efficiency~\cite{javaheripi2019swnet} has gained much research traction recently. Several works focus on designing automated methodologies for achieving such compact and accurate neural networks. GA has been applied to Neural Architecture Search (NAS)~\cite{xie2017genetic,real2017large,huang2018data,wang2019evolving,lu2019nsga,liang2019evolutionary} where the goal is to build a neural network architecture from scratch.
Different from NAS, this paper focuses on learning hyperparameters for DNN customization, which acts on pre-trained DNNs and does not require training-from-scratch. This property allows our algorithm to enjoy a fast search-space exploration using parallel computing resources.

\noindent\textbf{Automated DNN Compression. }Authors of~\cite{hu2018novel} develop a pruning scheme that selects the pruned filters using GA rather than magnitude-based or gradient-based approaches~\cite{li2016pruning,molchanov2016pruning}. Nevertheless, the amount of pruning applied per layer is still unknown. \sys{} addresses this remaining challenge by learning the compression hyperparameters. As a result, \sys{} can be applied to generalized compression techniques and not just network pruning.
Reinforcement Learning (RL)~\cite{he2018amc,wang2018haq} is another promising tool for DNN compression. Although effective in finding near-optimal solutions, 
RL applies compression one layer at a time to overcome the curse of dimensionality. Thus, the RL agent typically needs many learning episodes to identify the inherent inter-layer correlations.

GA is shown to produce competitive results with RL on the challenging tasks of neural architecture search and robotic control while enjoying higher scalability and significantly lower training overhead~\cite{salimans2017evolution,real2017large,xie2017genetic}. Inspired by this, we develop an optimization method based on GAs that achieves better/similar final solutions as prior art with much lower runtime and computational overhead. \sys{} optimizes all hyperparameters at once with a high scalability by exploiting parallelism specially in distributed settings.

%% file: 3_formulation.tex
\vspace{-0.7cm}
\section{Problem Formulation}\label{sec:formulation} 
DNN compression, in high-level, is a transformation $\mathcal T(M, \vec{x})$ that converts a pre-trained model $M$ to a compressed model $\hat M_{\vec{x}}$ with lower computational complexity. In this process, the adjustable hyperparameter vector $\vec x\in\mathbb{R}^d$ controls the complexity and accuracy of the output model.
A desirable compressed network $\hat M_{\vec{x}}$ should satisfy two properties:
(i)~the generalization capability of the compressed model should resemble the original network and 
(ii)~the execution cost of $\hat M_{\vec{x}}$ on the target hardware platform should be as low as possible. 
These properties outline a multi-objective black-box optimization problem over the hyperparameter vector $\vec x$, dubbed an \textit{individual}. We assume we have access to a scoring oracle, $f(\cdot)$, that assesses each individual based on the classification accuracy on a validation set and computational complexity of its corresponding compressed model. The scoring scheme is designed to capture the trade-off between model accuracy $A(\hat M_{\vec{x}})$ and hardware cost $C(\hat M_{\vec{x}}$). \sys{}'s customized score translates the original bi-objective problem of DNN compression to a single-objective optimization:

\begin{equation}\label{eq:overall_score}
\setlength{\abovedisplayskip}{0pt}
\setlength{\belowdisplayskip}{-2pt}
\underset{\vec x\in \mathbb{R}^d}\max{\ f(A(\hat M_{\vec{x}}), C(\hat M_{\vec{x}})),}
\end{equation}

\begin{figure*}[t]
    \centering
    \includegraphics[width=.9\textwidth]{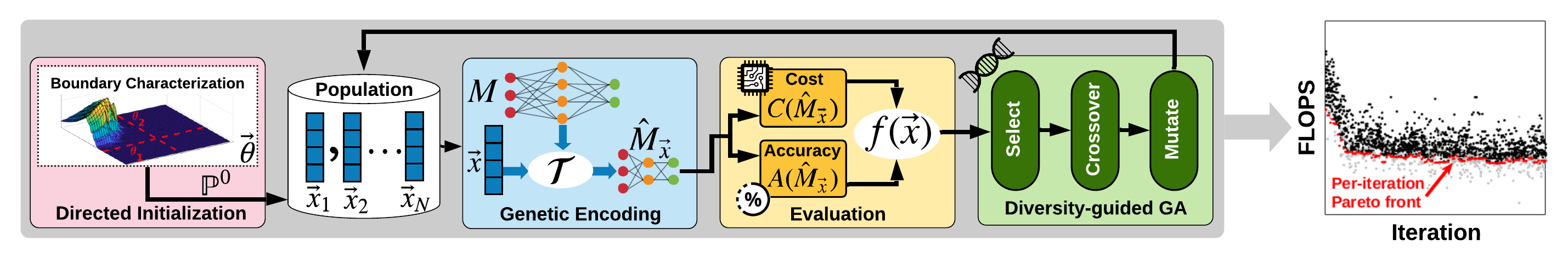}
    \vspace{-0.5cm}
    \caption{Overview of \sys{} genetic solution for hyperparameter customization. \sys{} iteratively improves the non-dominated front by lowering the FLOPs while maintaining a desired accuracy threshold.}
    \label{fig:overall}
    \vspace{-0.3cm}
\end{figure*}

For simplicity, we show $f(A(\hat M_{\vec{x}}), C(\hat M_{\vec{x}}))$ as $f(\vec{x})$ in the rest of the paper. Since full knowledge about $f(\cdot)$ and/or its first derivatives cannot be assumed, often empirical evaluations and optimization is the only viable strategy. Brute-force evaluation, in general, is infeasible as the search-space grows exponentially with $d$. Instead, we propose an empirical zero$^{th}$ order optimization algorithm based on
GA, dubbed \sys{}, for finding the maximum value in \eq{\ref{eq:overall_score}}, i.e., $f^*$. We denote the population at the $t^{th}$ step of the GA by $\mathbb{P}^t=\{\vec x_n^t\}_{n=1}^N$ where $N$ is the population size. \sys{} aims at finding a near-optimal population after $T$ iterations, $\mathbb{P}^T=\{\vec x^{T}_n\}_{n=1}^N$, where the individuals are in vicinity of the optimal solution:
\begin{equation}
     f(\vec x^{T}_n)\approx f^* \qquad \forall{n \in \{1,\dots,N\}}
\end{equation}

In particular, by adopting a guided search through genetic operations, \sys{} iteratively samples better DNNs with higher $f(\vec x)$.

%% file: 4_approach.tex
\section{\sys{} Approach}\label{sec:approach}
We provide a generic solution to effectively compress a pre-trained DNN while maximally preserving model accuracy. \sys{} automation policy acts on a pool of hyperparameters and explores the corresponding search-space using customized genetic operations. An overview of \sys{} genetic solution is shown in Figure~\ref{fig:overall} and Algorithm~\ref{alg:genetic} presents the corresponding pseudo code. Below, we summarize the key steps in our optimization framework: 

\begin{enumerate}[label=\Roman*., wide, labelindent=0pt, labelwidth=!, itemsep=3pt]
    \item First, a pre-processing step characterizes the search-space boundaries within which the optimal solution can reside. These boundaries are specified based upon task-enforced constraints on inference accuracy. Using the acquired boundaries, an initial population of hyperparameter vectors $\mathbb{P}^0=\{\vec x_1^0, \dots, \vec x_N^0\}$ are sampled (Sec.~\ref{sec:init}).
    \item At each iteration, all individuals $\vec x$ are translated to their compressed DNNs $\hat M_{\vec{x}}$~(Sec.~\ref{sec:translation}). The optimization scores $f(\vec x)$ are then evaluated in parallel (Sec.~\ref{sec:gen_eval}).
    \item Based on the new evaluations, \sys{} genetic operations, i.e., selection, crossover, and mutation (Sec.~\ref{sec:gen_ops}), are performed to update the population towards a new, more competent, generation.
\end{enumerate}

By iteratively applying steps II and III above, the non-dominated Pareto front for compressing a desired pre-trained DNN is obtained. In the following, we elaborate on each design component in detail.

\setlength{\textfloatsep}{0pt}
\begin{algorithm}[h]
\caption{\sys{} Search Algorithm} \label{alg:genetic}
\begin{flushleft}
\textbf{Inputs:} fitness oracle $f(\cdot)$, population size $(N)$, iterations $(T)$, crossover parameters $(P_{cross},P_{swap})$, mutation parameter $P_{tweak}$.

\textbf{Output:} Population after $T$ iterations, $\mathbb P^T=\{\vec x_n^T\}_{n=1}^N$. \\
\vspace{-0.2cm}
\hrulefill
\end{flushleft}
\begin{algorithmic}[1]
\State $\mathbb{P}^0 = \{\vec x_n^0\}_{n=1}^N$\Comment{\hlblue{Directed Initialization}}
\State $div_{thr}=\frac{div(\mathbb P^0)}{2}$
\For {$t = 0\dots T-1$}
\State  Calculate scores: $\{f_n = f(\vec x^{t}_n)\}_{n=1}^N$\Comment{\hlblue{Evaluation}}
\State $\mathbb{S}^{t}= sample(\mathbb{P}^{t}| \{f_n\}_{n=1}^N)$\Comment{\hlblue{Selection}}
\State Derive list of parents: $\mathbb S^t\rightarrow\mathbb{\hat S}^t=\{(p_1^n, p_2^n)\}_{n=1}^\frac{N}{2}$
\For{$(p_1, p_2)$ in $\mathbb{\hat S}^t$} \Comment{\hlblue{Crossover}}
\State $\mathbb C^t$.append(crossover$(p_1, p_2|P_{cross},P_{swap})$)
\EndFor
\If {$div(\mathbb{C}^t)\leq div_{thr}$}\Comment{\hlblue{Mutate}}
\State Adjust $P_{mutate}$
\State $\mathbb P^{t+1}$ = Mutate$(\mathbb C^t|P_{mutate},P_{tweak})$
\EndIf
\EndFor 
\end{algorithmic}
\end{algorithm}

\input{4_1_translation}
\input{4_2_evaluation}
\input{4_3_operations}
\input{4_4_initialization}

%% file: 4_1_translation.tex
\vspace{-0.7cm}
\subsection{Genetic Encoding}\label{sec:translation}
An initial step for application of \sys{} is defining the pertinent search-space for the black-box optimization. To this end, we are in need of a global encoding scheme $\mathcal{T}(M, \vec x)$ that translates a vector of compression hyperparameters $\vec x$ to its corresponding compressed architecture $\hat M_{\vec x}$. To ensure an effective search, our proposed translation possesses the following characteristics: 
\begin{itemize}[leftmargin=*, itemsep=0pt]
    \item \textbf{Distinctness.} Our encoding is bijective: given an individual $\vec x$ and model $M$, the compressed $\hat{M}_{\vec x}$ is uniquely determined.
    \item \textbf{Continuity.} A small change in an individual translates to a similarly small alternation in the corresponding compressed DNN. This is particularly important to ensure search convergence.
    \item \textbf{Scalability.} \sys{} encoding offers a compact, low dimensional representation, enabling effective search. The search-space dimensionality scales linearly with the number of DNN layers and is invariant to number of DNN parameters/operations.
    \vspace{-0.2cm}
\end{itemize}

\begin{figure}[b]
    \centering
    \includegraphics[width=0.8\columnwidth]{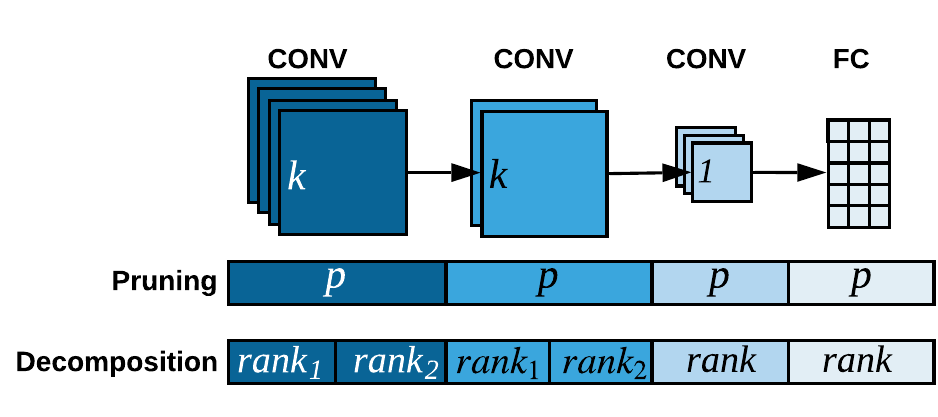}
    \vspace{-0.5cm}
    \caption{Vectorized representation of an example $4$-layer DNN for \textit{Pruning} and \textit{Decomposition}. Here, CONV and FC denote convolutional and fully-connected layers, respectively.
    }\label{fig:individuals}
\end{figure}

We direct our focus on \textit{four} compression tasks, namely, structured~\cite{he2017channel} and non-structured~\cite{han2015learning} Pruning, Singular Value Decomposition \cite{zhou2016less}, and Tucker-2 approximation~\cite{kim2015compression}.
Figure~\ref{fig:individuals} shows  a high-level view of our encoding scheme for a $4$-layer neural network. 
For pruning, we allocate $1$ hyperparameter per layer: $\vec x\in\mathbb{R}^4$. For low-rank approximation, we allocate $2$ hyperparameters for Tucker-2 and $1$ hyperparameter for SVD, per layer, resulting in $\vec x\in\mathbb{R}^6$. In the following, we explain each method's hyperparameters and their interpretation as a compressed model in detail.

\noindent{\bf Pruning.}  
Pruning reduces model size by setting a percentage of low priority parameters/channels to zero~\cite{li2016pruning,he2017channel,han2015learning}. 
We allocate one continuous value $p\in[0, 1]$, per layer, to represent the ratio of non-zero values. For the example of ResNet-50, each sample will thus be a vector $\vec x\in[0,1]^{50}$. We consider two contemporary DNN pruning methods, namely, structured and non-structured pruning. Structured pruning aims at removing a portion of feature-map channels while non-structured pruning removes a subset of DNN layer weights.

Having defined the vector of hyperparameters $\vec x$ for pruning, we implement the transformation $\mathcal{T}(M,\vec x)$ following common practice in prior art. For structured pruning, we use the sum of absolute gradients of model loss with respect to ReLU feature-map channels for pruning priority~\cite{molchanov2016pruning}.
For a ReLU layer with $c$ feature-map channels and pruning rate $p$, the $\lfloor p \times c\rfloor$ channels with lowest priorities (lower absolute gradients) are removed.
For non-structured pruning, we use the absolute value of weight parameters to prioritize them~\cite{han2015learning}.
Given a weight $W\in \mathbb{R}^{k\times k\times c\times f}$ and pruning rate $p$, the $\lfloor{p\times k\times k\times c\times f\rfloor}$ elements with lowest absolute values are pruned.

\vspace{0.1cm}
\noindent{\bf SVD.} We apply SVD on weight parameters ($W$) of fully-connected layers ($W \in \mathbb{R}^{c\times f}$) and point-wise convolutions ($W \in \mathbb{R}^{1\times 1 \times c \times f}$). 
For SVD, the compression parameter is the decomposition rank which takes an integer value in $\{1,\dots, R\}$ where $R=min(c,f)$. For efficiency, we discretize the space of approximation ranks in each layer into $64$ values and encode them as follows:
\begin{equation}
    \resizebox{0.8\columnwidth}{!}{
        $rank \in \{1\gets\frac{R}{64},\ \dots,\ 64\gets R\},\ \  R = min\{c,f\}$
    }\label{eq:quant_svd}
\end{equation}

\noindent{\bf Tucker-2.} Tucker decomposition is a generalized Higher Order SVD (HOSVD) for arbitrary-shaped tensors. We apply this method on 4-way weight tensors in convolutional layers, $W\in \mathbb{R}^{k\times k \times c\times f}$. We focus on Tucker-2 which only decomposes the tensor along $c$ and $f$ directions, i.e., output and input channels. 
For Tucker-2 decomposition, the compression hyperparameter for each convolution layer is a tuple of integer-valued approximation ranks $(r_1,r_2)$ where $r_1 \in \{1,\dots, c\}$ and $r_2 \in \{1,\dots, f\}$.
To increase search efficiency, we quantize the space of decomposition ranks to $8$ bins per-way as follows:

\begin{equation}
\setlength{\abovedisplayshortskip}{-5pt}
\setlength{\belowdisplayshortskip}{-5pt}
\setlength{\abovedisplayskip}{-5pt}
\setlength{\belowdisplayskip}{-5pt}
\resizebox{0.9\columnwidth}{!}{
    $r_1\in\{1\gets\frac{c}{8},\ \dots,\ 8\gets c\} \qquad
    r_2\in\{1\gets\frac{f}{8},\ \dots,\ 8\gets f\}$
}\label{eq:quant_tucker}
\end{equation}

%% file: 4_2_evaluation.tex
\subsection{Objective Evaluation}\label{sec:gen_eval}
\begin{figure}[h]
    \centering
    \includegraphics[width=0.75\columnwidth]{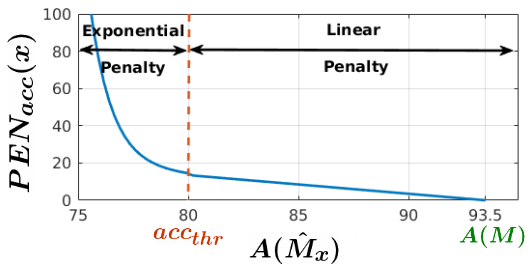}
    \vspace{-0.4cm}
    \caption{\sys{} accuracy penalty with $acc_{thr}=80\%$ when original model accuracy is $A(M)=93.5\%$.}
    \label{fig:acc_penalty}
\end{figure}

We develop a customized scoring mechanism to assess the quality of individuals in each iteration of \sys{} algorithm. The multi-objectvie score of \sys{} simultaneously reflects the compressed DNN's accuracy and computational complexity. The objective of DNN compression then translates to maximizing this score. We formalize DNN compression as a constrained optimization as follows:
\begin{equation}\label{eq:constrained}
\setlength{\abovedisplayskip}{0pt}
\setlength{\belowdisplayskip}{0pt}
\setlength{\abovedisplayshortskip}{0pt}
\setlength{\belowdisplayshortskip}{0pt}
\begin{tabular}{l}
    $\underset{\vec x \in \mathbb{R}^d}{\max}\  \Delta C(M, \vec x)$\\ 
    $subject\ to\ \ A(\hat M_{\vec x}) > acc_{thr}$
\end{tabular}
\end{equation}
where $\Delta \mli{C}(M, \vec x)$ represents the normalized difference in hardware cost, e.g., number of floating-point operations (FLOPs), between the uncompressed network, $M$, and the compressed model, $\hat M_{\vec x}$. The function $A(\cdot)$ denotes the inference accuracy and $acc_{thr}$ is a task-enforced threshold on the post-compression accuracy. Having an accuracy constraint is crucial since the optimization algorithm will converge to a model size of zero otherwise. To solve the constrained optimization problem in \eq{\ref{eq:constrained}}, we formulate it as the following primal unconstrained optimization using penalty methods~\cite{back1997handbook}:
\begin{equation}\label{eq:logscore}
\setlength{\abovedisplayskip}{3pt}
\setlength{\belowdisplayskip}{3pt}
    \underset{\vec x \in \mathbb{R}^d}{\max}\  \Delta C(M,\vec x) - log(\mli{PEN}_{acc}(\vec x)))
\end{equation}
where the term $log(\mli{PEN}_{acc}(\vec x))$ is the exterior penalty function~\cite{chong2013introduction} that enforces a constraint on the accuracy, i.e., $A(\hat M_{\vec x})>acc_{thr}$. The function $\mli{PEN}_{acc}(\vec x)$ measures the accuracy degradation as follows:
\begin{equation}\label{eq:acc_penalty}
\setlength{\abovedisplayskip}{5pt}
\setlength{\belowdisplayskip}{5pt}
\resizebox{0.9\columnwidth}{!}{
 $\mli{PEN}_{acc}(\vec x) = \begin{cases}
 A(M) - A(\hat M_{\vec x}) \qquad \qquad \qquad \quad \ \ \ A(\hat M_{\vec x})\geq acc_{thr}\\
 A(M) - A(\hat M_{\vec x}) + e^{acc_{thr} - A(\hat M_{\vec x})} \quad A(\hat M_{\vec x})<acc_{thr}
 \end{cases}$}
\end{equation}
Figure~\ref{fig:acc_penalty} visualizes the accuracy penalty. To prevent undesirable drop of accuracy, we greatly diminish the score of individuals that cause lower accuracies than the set constraint, $acc_{thr}$. The $log$ penalty term can thus be estimated as follows:

\begin{equation}\label{eq:log_acc_penalty}
\setlength{\abovedisplayskip}{0pt}
\setlength{\belowdisplayskip}{0pt}
\resizebox{0.9\columnwidth}{!}{
 $log(\mli{PEN}_{acc}(\vec x))) = \begin{cases}
 log(A(M) - A(\hat M_{\vec x})) \quad \ A(\hat M_{\vec x})\geq acc_{thr}\\
 acc_{thr} - A(\hat M_{\vec x}) \qquad \quad A(\hat M_{\vec x})<acc_{thr}
 \end{cases}$}
\end{equation}

For accuracy values satisfying the threshold, this term enforces the accuracy maximization objective and applying the logarithm smoothens the accuracy variations by damping sudden changes. For accuracy values below $acc_{thr}$, a linear penalty is applied to prevent further accuracy loss.
To prevent numerical instability, we define our score function as the exponential of the primal optimization of \eq{\ref{eq:logscore}} which is formalized in \eq{\ref{eq:score}}:

\begin{equation}\label{eq:score}
\setlength{\abovedisplayskip}{0pt}
\setlength{\belowdisplayskip}{0pt}
\setlength{\abovedisplayshortskip}{0pt}
\setlength{\belowdisplayshortskip}{0pt}
\resizebox{0.35\columnwidth}{!}{
    $f(\vec x)$ = $\frac{e^{\Delta \mli{C}(M, \vec x)}}{\mli{PEN}_{acc}(\vec x)}$
}
\end{equation}

Note that maximizing the score function of \eq{\ref{eq:score}} is equivalent to maximizing its logarithm value in \eq{\ref{eq:logscore}}.
The developed score function encourages reduction in hardware cost (in the numerator) while penalizing the decrease in model accuracy caused by compression (in the denominator). 
Our proposed scoring mechanism is globally applicable to various compression tasks and can be modified to reflect a variety of hardware costs, e.g., power, memory footprint, and runtime. 
To ensure efficiency, inference accuracies are measured on a small held-out portion of the training samples. \sys{} scoring function successfully models the ultimate goal of high compression with minimal accuracy degradation.

%% file: 4_3_operations.tex
\vspace{-0.1cm}
\subsection{Optimization with Genetic Operations}\label{sec:gen_ops}

\sys{} genetic optimization is a metaheuristic approach inspired by natural evolution and the notion of ``survival of the fittest'' and can be leveraged to explore large search-spaces. Our GA works on a \textit{population} of \textit{individual}s.
The core idea is to encourage creation of superior individuals and elimination of the inferior ones. To this end, an iterative process evolves the previous generation into a new, more competent population by performing a set of bio-inspired actions, i.e., \textit{selection}, \textit{cross-over}, and \textit{mutation}. 

\begin{figure}[h]
    \centering
    \vspace{-0.1cm}
    \includegraphics[width=1.0\columnwidth]{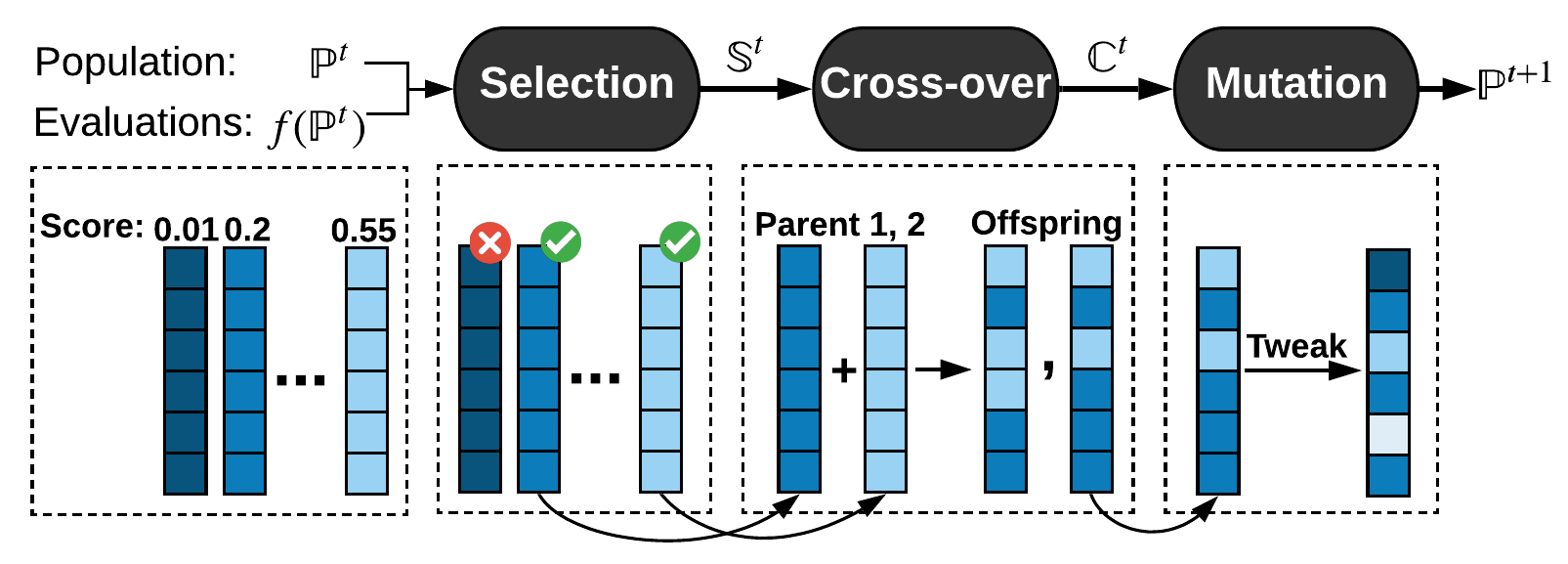}
    \vspace{-0.7cm}
    \caption{Genetic operations for creating a new population.}
    \label{fig:genetic_ops}
\end{figure}

Figure~\ref{fig:genetic_ops} illustrates the process of evolving a new population from the previous one. After the current population is evaluated (see \eq{\ref{eq:score}}), individuals are assigned scores representing their quality, i.e., \textit{fitness}. The selection step then performs a sampling (with replacement) based on individuals' fitness scores. 
Cross-over and mutation aim to explore the proximity of the selected individuals by injecting small random patterns. 
To promote a diverse non-dominated front of solutions, we tune the hyperparmeters associated with mutation based on a diversity measure. This, in turn, maintains the balance between \textit{exploration} and \textit{exploitation}.

\sys{} genetic operations are devised to be compatible with the translation method. In particular, our designed operations satisfy the ``closure'' property: the output vector of mutation/cross-over translates to a valid compressed DNN. Note that selection, and cross-over are unified across different compression techniques while the mutation operation needs to be customized per compression method.

\noindent\textbf{Selection. } 
\sys{} selection stage attempts to choose high-quality individuals based on the ordering of Pareto dominance to generate the next population. 
\sys{} adopts a fitness-proportionate selection mechanism. Let us denote the population at the beginning of $t$-th iteration by $\mathbb P^{t}=\{\vec x_{n}^t\}_{n=1}^N$ with fitness scores $\{f_n\}_{n=1}^N$ obtained from \eq{\ref{eq:score}}. To select superior samples, we construct the following probability distribution over the population using the fitness scores:
\begin{equation}\label{eq:norm_score}
    p_n = \frac{f_n-f_{min}}{\sum_{n=1}^{N} (f_n-f_{min})},
\end{equation}
where $f_{min}$ is the minimum score. Subtraction of the minimum score ensures that the probability of selecting the weakest individual is zero and it is always eliminated. The superior individuals, $\mathbb S^{t}$, are selected by performing a non-uniform random sampling of the old population where the probability of selecting an individual is derived from \eq{\ref{eq:norm_score}}. 
The total number of individuals in the population remains unchanged across iterations. Superior individuals can be selected multiple times for the new population (sampling with replacement) while ones with low scores are rarely chosen. The combination of \sys{} evaluation and selection stages preserves high-quality individuals and eliminates weak ones. 

\noindent\textbf{Crossover. }
We design a crossover operation which creates new individuals, i.e., new compression configurations, $\mathbb{C}^t$ by inheriting and combining compression rates from a pair of parent individuals. The crossover is performed by randomly swapping corresponding elements of parent genotypes.
Given the selected population $\mathbb{S}_{t}$, we sort the individuals in descending order based on their fitness score. To form the cross-over pairs, we pick the fittest individual available as the first parent. We then choose the individual which has highest \textit{distance} with the first parent as the second parent. Such distance-based selection of pairs is motivated by increasing diversity among the newly generated offspring, and promoting exploration. For two individuals $\vec x_1, \vec x_2\in\mathbb R^d$, the distance is calculated as follows:
\begin{equation}
    dist(\vec x_1, \vec x_2) = \frac{1}{d\times K}~\sqrt{\sum_{i=1}^{d} (\vec x_1[i] - \vec x_2[i])^2}
\end{equation}
where $K$ is a constant equal to the maximum allowed value for the corresponding compression hyperparameter, e.g., $K=64$ for SVD decomposition. We use two parameters to control the crossover operation: $p_{cross}$ determines the probability of applying crossover between the parents, and $p_{swap}$ is the per-element swapping probability. The proposed crossover allows superior individuals to exchange learned patterns and transfer knowledge across the population.

\noindent\textbf{Mutation.}
Mutation randomly tweaks each individual in the crossovered population $\mathbb C^t$ to create a new population $\mathbb P^{t+1}$ for the next iteration of the GA. We devise two variations of mutation for continuous and discrete-valued individuals.
\begin{itemize}[leftmargin=*]
    \item \textbf{Continuous-valued Individuals.} 
    Each element of a continuous-valued individual is mutated by adding a random value drawn from a zero-mean Normal distribution $\mathcal{N}(0,\,0.2)$. We then clip the values to ensure they remain in the allowed interval, i.e., $[0,1]$. 
    \item \textbf{Discrete-valued Individuals.} 
Discrete individuals are mutated by randomly incrementing/decrementing vector elements such that the value remains in the valid discrete range in Sec.~\ref{sec:translation}. 
\end{itemize}

Mutation allows exploration of the neighborhood of candidate points in the search-space. Similar to cross-over, we define two control parameters: $P_{mutate}$ is the probability that the individual gets mutated and $P_{tweak}$ determines the per-element tweaking probability. To maintain the balance between exploration and exploitation and avoid premature convergence, we tune the mutation parameters based on the diversity of individuals in $\mathbb C^t$. We use the dispersion of individuals in the population and define diversity as follows:
\begin{equation}
    div(\mathbb C^t) = ~\frac{1}{N}\sum_{n=1}^N \left[\sum_{i=1}^d (\vec x_n^t[i] - \vec\mu_{x}^t[i])^2\right]
\end{equation}
where $\vec\mu_x^t \in\mathbb R^d$ is the mean of all individuals in $\mathbb C^t$. As can be seen, the diversity function is closely tied to the variance. The diversity can therefore be adjusted by controlling the per-element variance of the individuals in the crossovered population $Var[X_i]$. Let us denote an arbitrary individual after crossover by $\vec x\in\mathbb C^t$, which transforms to $\vec y\in\mathbb P^{t+1}$ after mutation. The $i^{th}$ element of $\vec y$ is thus sampled from a random variable $Y_i$ with the following probability distribution: 
\begin{equation}\label{eq:Y_mutate}
\resizebox{0.7\columnwidth}{!}{
$P(Y_i=\vec y[i])=\begin{cases} p_M \qquad\quad \ \ \vec y[i]=\vec x[i]+\eta \\
1-p_M \qquad \vec y[i]=\vec x[i]
 \end{cases}$}
\end{equation}
where $P_M=P_{mutate}\times P_{tweak}$ and $\eta$ is the random perturbation applied during mutation. Note that $\vec x$ is also a random variable. The per-element variance of $\vec y$ is thus:
\begin{equation}\label{eq:var_mutate}
    Var[Y_i]=Var[X_i] + p_M \sigma_\eta^2
\end{equation}
Here, $\sigma_\eta^2$ is the variance of the added perturbation. Summing up the vector values in \eq{\ref{eq:var_mutate}} provides the new population diversity:
\begin{equation}\label{eq:div}
    div(\mathbb P^{t+1}) = div(\mathbb C^t)+d\times P_M\sigma_\eta^2
\end{equation}
For a desired threshold on population diversity, we can therefore determine the mutation parameters using \eq{\ref{eq:div}}. Since tweaking multiple elements of the individual vector can result in drastic changes in the corresponding compressed DNN's architecture and accuracy, we restrict $P_{tweak}$ to a small value ($P_{tweak}=0.05$) and merely adjust $P_{mutate}$ for diversity control. In our experiments, we set the diversity threshold to be half the diversity for the randomly initialized population. Such adaptive tuning of mutation allows for a diversity-guided search and ensures a fast and
stable convergence.

%% file: 4_4_initialization.tex
\vspace{-0.2cm}
\subsection{Directed Initialization}\label{sec:init}
\begin{figure}[h]
    \centering
\includegraphics[width=0.77\columnwidth]{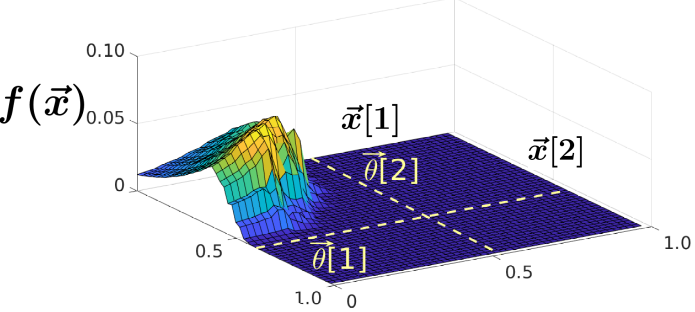}
    \vspace{-0.4cm}
    \caption{Search space for a two-layer network pruning.}
    \label{fig:mesh_grids}
\end{figure}

A na\"ive random initialization of samples in the first iteration can result in a slow and sub-optimal convergence. To address this, we utilize boundary characterization as a pre-processing step to enable a targeted sample initialization.
The general behavior of inference accuracy for a compressed DNN $\hat{M}_{\vec x}$ is monotonic: as the compression rates increase, the accuracy drops. 
As such, for any task-enforced threshold on accuracy, we can characterize the boundaries of $\vec x[i]$ on a per-layer basis. Boundary Characterization allows for a directed search that eliminates
unnecessary exploration of outlier subspaces, i.e., regions that are unlikely to contain the optimization solution. Figure~\ref{fig:mesh_grids} visualizes the hyperparameter search-space and the outlier regions for pruning a two-layer neural network. The horizontal plane corresponds to different configurations of per-layer pruning rates and the vertical axis represents their quality  (score $f(.)$).

The optimal solution to the search problem is a configuration with the highest score. 
The outlier regions in Figure~\ref{fig:mesh_grids} are therefore the flat sectors of the space. 
By filtering out the non-optimal outlier regions, \sys{} sampling can find the solution within the region bounded by the initialization margins.
We find a threshold vector $\vec \theta$ where each element constrains a single hyperparameter that corresponds to a certain layer in the compressed DNN. In the example of Figure~\ref{fig:mesh_grids}, $\vec \theta$ has two elements, each presented by a dashed line ($\vec\theta[1]$ and $\vec\theta[2]$).
Below we describe how the boundaries $\theta[i]$ are obtained given an accuracy threshold $acc_{thr}$ for each compression task.

\vspace{0.1cm}
\noindent{\bf Pruning.} We obtain a threshold vector $\vec\theta \in \mathbb{R}^{d}$ with the $i^{th}$ element $\vec\theta[i]$ specifying the maximum pruning rate of the $i^{th}$ layer such that the inference accuracy of the compressed model $\hat M_{\vec x}$ does not violate the accuracy threshold $acc_{thr}$:
\begin{equation}\label{eq:prune_th_init}
\resizebox{0.88\columnwidth}{!}{
$\vec\theta[i] = max\{p\}\ \ s.t. \ \ \vec x[j]=\begin{cases} 
      p & j = i \\
      0 & j \neq i 
   \end{cases}\  \&\ \  A(\hat{M}_{\vec x})>acc_{thr}$
}
\end{equation}

\vspace{0.1cm}
\noindent{\bf Decomposition.} For SVD and Tucker-2 decompositions, the threshold vector $\theta$ represents per-layer minimum ranks satisfying $acc_{thr}$:
\begin{equation}
\resizebox{0.90\columnwidth}{!}{
$\vec\theta[i] = min\{r\}\ \ s.t. \ \ \vec x[j]=\begin{cases} 
      r & j = i \\
      K_j & j \neq i 
  \end{cases}\  \&\ \  A(\hat M_{\vec x})>acc_{thr}$
}
\end{equation}
where $K_j\in\{64, 8\}$ is the maximum allowed encoded rank for the $j^{th}$ hyperparameter (see equations ~\ref{eq:quant_svd}\&\ref{eq:quant_tucker}).

Note that real-world neural networks have many more layers and the pertinent search-space is of much higher dimensionality than the $2$-dimensional plane shown Figure~\ref{fig:mesh_grids}. For a $d$-dimensional search-space, the proposed boundary characterization scheme reduces the effective (continuous) search volume from $1$ to $\prod_{i=1}^d \vec\theta[i]$ for pruning. For decomposition, the (discrete) search-space size is reduced from $\prod_{i=1}^d K_i$ to $\prod_{i=1}^d (K_i-\vec\theta[i])$. Such approach significantly improves search convergence time and solution quality.

After performing boundary characterization, we randomly sample the initial population from the space enclosed by the threshold vector $\vec \theta$. For pruning, the $i^{th}$ element $\vec x[i]$ in each sample vector is drawn from a Normal distribution $\mathcal{N}(\vec\theta[i]/2,\vec\theta[i]/2)$. For decomposition, the $i^{th}$ element $\vec x[i]$ is randomly selected from $\{\vec \theta[i], \vec \theta[i]+1, \dots, K_i\}$.

%% file: 5_experiments.tex
\vspace{-0.5cm}
\section{Experiments}\label{sec:experiments}
We provide extensive evaluations on CIFAR-10~\cite{krizhevsky2009learning} and ImageNet~\cite{russakovsky2015imagenet} benchmarks. The evaluated network architectures include AlexNet, VGG, ResNet family, and MobileNets. 
All models are implemented using Pytorch library.
The networks are trained from scratch following the parameter setup and training schedule adopted by the original papers~\cite{he2016deep,simonyan2014very,krizhevsky2012imagenet,howard2017mobilenets,sandler2018mobilenetv2}.
For CIFAR-10, we use a VGG-variant as used in~\cite{jiang2018efficient}. We compress the models using \sys{} and fine-tune them for $20$ and $60$ epochs on CIFAR-10 and ImageNet, respectively.
The genetic parameters are set as $P_{tweak}=0.05$, $P_{cross}=0.8$, $P_{swap}=0.2$~\cite{xie2017genetic}. $P_{mutate}$ is adaptively determined based on the population diversity, as discussed in Sec.~\ref{sec:gen_ops}. All results are from a single run of the GA.

\vspace{-0.1cm}
\input{5_1_cifar}
\vspace{-0.1cm}
\input{5_2_imagenet}
\vspace{-0.3cm}
\input{5_3_mobilenets}

\subsection{Search Overhead and Scalability}
The core computational load in \sys{} algorithm corresponds to the evaluation of a batch of $b$ samples. For each sample $\vec x_i$ in the batch, the evaluation phase comprises transforming the sample to its corresponding compressed DNN $\hat{M}_{\vec x_i}$, measuring inference accuracy on the validation data, and emulating the execution cost.

Since samples in a batch are independent, evaluation can be well-parallelized on multiple GPU devices to achieve faster search convergence. Table~\ref{tab:runtime} summarizes the runtime of \sys{} algorithm for several benchmarks and datasets. Runtimes are measured on a machine with an Intel Xeon E5 CPU and four NVIDIA Titan Xp GPUs. The results show high scalability: runtime drops almost linearly with the number of GPUs. The state-of-the-art RL algorithm reports $\sim1$ hour to compress CIFAR-10 architectures~\cite{he2018amc}. For their most complex benchmark, i.e., ResNet-56, \sys{} achieves a search time of only $\sim 12$ minutes on a single GPU and $\sim 3$ minutes on four GPUs.

\begin{table}[h]
\centering
\vspace{-0.3cm}
\caption{Search runtime of \sys{} for pruning on various benchmarks. Here, $b$ denotes the number of samples per iteration and $N_{iters}$ is the number of search iterations.}\label{tab:runtime}
\vspace{-0.3cm}
\resizebox{1.0\columnwidth}{!}{
\begin{tabular}{clcccccc}
\hline
\multirow{2}{*}{Dataset}                      & \multirow{2}{*}{Arch.} & \multirow{2}{*}{$b$} & \multirow{2}{*}{$N_{iters}$} & \multicolumn{4}{c}{Search Time (minutes)} \\ \cline{5-8} 
                                              &                        &                    &                           & 1 GPU    & 2 GPU    & 3 GPU    & 4 GPU    \\ \hline
\multirow{5}{*}{\rotatebox[origin=c]{90}{ImageNet}} & AlexNet                & 50                 & 50                        & 10       & 5        & 3        & 3        \\
\multicolumn{1}{l}{}                          & VGG-16                 & 50                 & 50                        & 112      & 57       & 38       & 28       \\
\multicolumn{1}{l}{}                          & ResNet-50              & 100                & 50                        & 145      & 73       & 49       & 36       \\
\multicolumn{1}{l}{}                          & MobileNetV1            & 50                 & 100                       & 97       & 48       & 32       & 24       \\
\multicolumn{1}{l}{}                          & MobileNetV2            & 50                 & 100                       & 116      & 57       & 38       & 25       \\ \hline
\multirow{4}{*}{\rotatebox[origin=c]{90}{CIFAR-10}}  & VGG                    & 50                 & 50                        & 3        & 2        & 1        & 1        \\
                                              & ResNet-50              & 100                & 50                        & 35       & 19       & 13       & 11       \\
                                              & ResNet-56              & 100                & 50                        & 12       & 7        & 5        & 3        \\
                                              & ResNet-110             & 200                & 50                        & 55       & 30       & 22       & 16       \\ \hline
\end{tabular}}
\end{table}

\setcounter{figure}{8}
\begin{figure*}[b]
    \centering
    \vspace{-0.2cm}
\includegraphics[width=0.9\textwidth]{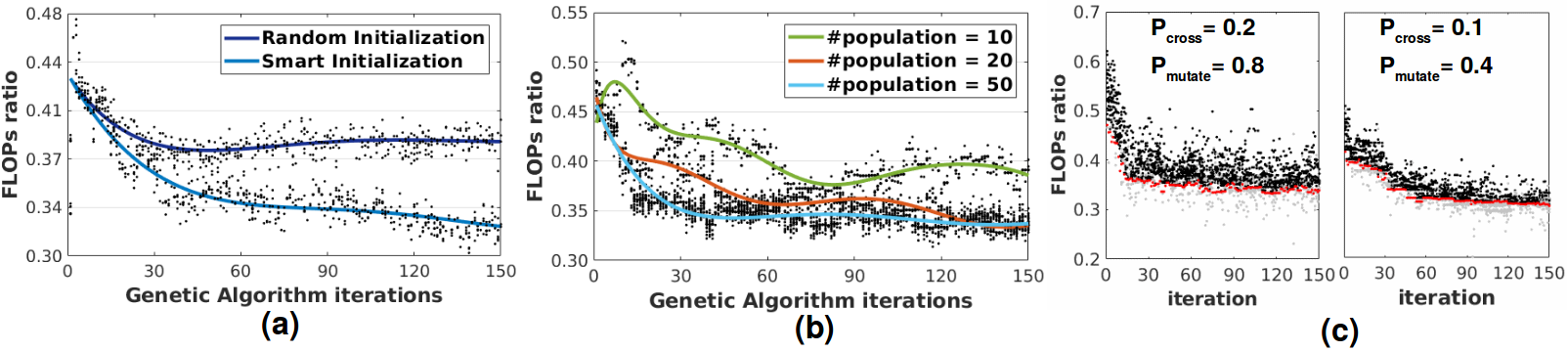}
    \vspace{-0.2cm}
    \caption{Ablation studies for VGG on CIFAR-10:(a) Effect of initialization method. (b) Effect of population size. (c) Effect of mutation and cross-over probabilities. Here, individuals shown in grey are those with lower accuracy than $acc_{thr}=60\%$ while individuals shown in black are those that meet the accuracy threshold. Red dots correspond to the highest-quality individual in that iteration.}
    \label{fig:ablation}
    \vspace{-0.2cm}
\end{figure*}
\setcounter{figure}{6}

\subsection{Analysis and Discussion}\label{sec:discussion}
\begin{figure}[h]
    \centering
    \vspace{-0.3cm}
\includegraphics[width=1\columnwidth]{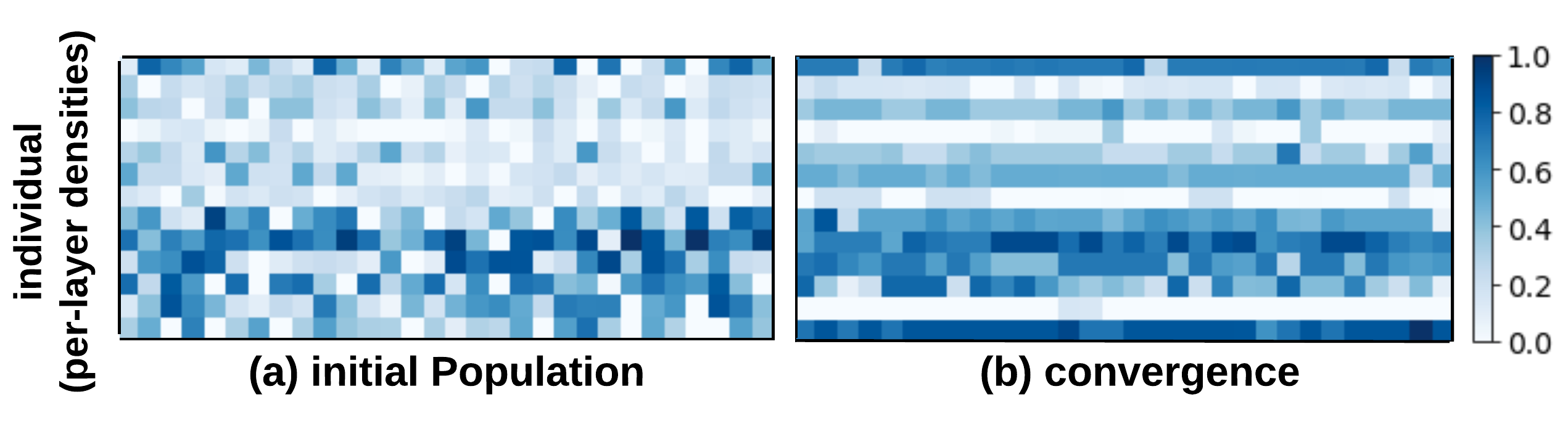}
    \vspace{-0.7cm}
    \caption{(a)~Initialized population for structured pruning. (b)~Population upon convergence.}
    \label{fig:heatmap}
\end{figure}

To illustrate \sys{} methodology, we consider VGG architecture trained on CIFAR-10 compressed with structured (filter) pruning for $acc_{thr}=60\%$. 
The initial samples obtained from our directed initialization scheme are shown in Figure~\ref{fig:heatmap}-a, where each column corresponds to an individual and each row represents a model layer.
After applying \sys{} for $50$ iterations, the population pattern of Figure~\ref{fig:heatmap}-b is learned. Upon convergence, individuals have similarly high scores. 
\sys{} successfully learns expert-designed rules: first and last layers of the network (first and last rows in Figure~\ref{fig:heatmap}-b) are given high densities to preserve accuracy.

\begin{figure}[t]
    \centering
\includegraphics[width=0.85\columnwidth]{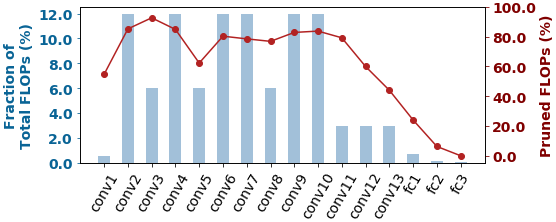}
    \vspace{-0.4cm}
    \caption{Per-layer FLOPs (bar charts) and percentage of pruned FLOPs (curve) for VGG16 trained on ImageNet. 
    \sys{} prunes more from layers with higher FLOPs.
    }
    \label{fig:perlayer_flops}
\end{figure}

\sys{} performs \textit{whole-network} compression by encoding all layers' hyperparameters in one genotype. As such, \sys{} effectively learns which hyperparameter configuration least affects model accuracy and most reduces the overall FLOPs. To show this capability, we present the per-layer FLOPs for VGG-16 network trained on ImageNet in Figure~\ref{fig:perlayer_flops}. The bars show the per-layer FLOPs percentage in the original model; the curve shows the per-layer percentage of pruned FLOPs in the compressed DNN. 
Different from prior art~\cite{jiang2018efficient}, \sys{} prunes the first convolutions more and relaxes pruning for later layers as they have a minor role in FLOPs. 

\subsection{Ablation Study}\label{sec:ablation}
\vspace{-0.1cm}
In this section, we study the effect of various \sys{} parameters on algorithm convergence and final FLOPs/accuracy. For brevity, we only focus on the task of structured pruning for VGG on CIFAR-10. We show the trend lines as well as a fraction of individuals (black dots) across \sys{} iterations.

\noindent\textbf{Effect of Initialization.} 
Figure~\ref{fig:ablation}-a shows the evolution of FLOPs ratio for two initialization policies, one with uniformly random samples and one with our proposed initialization scheme discussed in Sec.~\ref{sec:init}. As seen, naive initialization greatly harms the convergence rate and final FLOPs.

\noindent\textbf{Effect of Population Size.}  Figure~\ref{fig:ablation}-b presents the effect of population size on \sys{} convergence. A higher number of individuals results in a smoother convergence and lower final FLOPs. This effect saturates for a large enough population. 
In our experiments, we observed that the population size should be larger for searching vector spaces with higher dimensionality.

\noindent\textbf{Effect of Mutation and Cross-over Parameters.} 
We conduct two experiments, one with $(P_{mutate}, P_{cross}) = (0.8, 0.2)$ and the other with $(0.4, 0.1)$ and compare the convergence in Figure~\ref{fig:ablation}-c. Higher $(P_{mutate}, P_{cross})$ allows the GA to explore more possibilities, leading to faster convergence while smaller probabilities result in a more stable evolution. Thus, \sys{} dynamically tunes the GA parameters to ensure balanced exploration and exploitation (see Sec.~\ref{sec:gen_ops}).

%% file: 5_1_cifar.tex
\vspace{-0.2cm}
\subsection{Quantitative Results on CIFAR-10}\label{sec:cifar-10}
We apply \sys{} to pre-trained CIFAR-10 architectures and compare our results with prior art in Table~\ref{tab:cifar-10}.
Specifically, we conduct experiments with non-structured pruning ($P_n$), structured pruning ($P_s$), SVD and Tucker decomposition ($D$), and combination of multiple compression methods ($D+P_s$).
We set the population size to $100$ for ResNet-56 and ResNet-50, $200$ for ResNet-110, and $50$ for VGG. We randomly select $1,000$ images from the training data to use as validation set for score computation. The threshold ($acc_{thr}$) is set to $90\%$ for ResNet-X and $60\%$ for VGG. For all networks, we let \sys{} run for $50$ iterations.

\vspace{0.1cm}
\noindent\textbf{Non-structured Pruning ($P_n$ in Table~\ref{tab:cifar-10}).} We perform non-structured pruning on ResNet-50 and report the ratio of non-zero model parameters. 
$acc_{thr}$, is set to $93\%$ and we do not perform any fine-tuning on the compressed model. As shown, \sys{} achieves $1.34\times$ lower parameters compared to state-of-the-art RL method~\cite{he2018amc}. 
Note that the lower FLOPs and comparable accuracy of~\cite{liu2018rethinking} are due to training the model from scratch whereas \sys{} and~\cite{he2018amc} do not include any fine-tuning. 

\vspace{0.1cm}
\noindent\textbf{Structured Pruning ($P_s$ in Table~\ref{tab:cifar-10}).} We implement structured pruning by adding masks after $ReLU$ activation layers. 
We base our comparisons on FLOPs per inference relative to the uncompressed baseline. With similar or higher accuracy, \sys{} compressed models achieve $1.35\times$ lower FLOPs than prior art (on average). 

\vspace{0.1cm}
\noindent\textbf{Decomposition and Pruning ($D+P_s$ in Table~\ref{tab:cifar-10}).} To unveil the full optimization potential of our method, we allow \sys{} to learn and combine multiple compression techniques, namely, pruning, SVD, and Tucker. The $D+P_s$ experiments are conducted by first decomposing the network and then applying structured pruning. As shown in Table~\ref{tab:cifar-10}, \sys{} pushes the limits of compression by $\mathbf{2.45\times}$ on average with less than $1\%$ drop in accuracy compared to state-of-the-art works. We also report FLOPs reduction by only applying decomposition (SVD and Tucker), shown by $D$ in Table~\ref{tab:cifar-10}. 

%% file: 5_2_imagenet.tex
\vspace{-0.5cm}
\subsection{Quantitative Results on ImageNet}\label{sec:imagenet}
Table~\ref{tab:imagenet} summarizes \sys{} results for ImageNet dataset.
Number of samples is $100$ for ResNet-50, $50$ for VGG-16, and $20$ for AlexNet. We let \sys{} run for $50$ iterations with $acc_{thr}=10\%$ for all models. This target accuracy is compensated by fine-tuning.

\noindent\textbf{Non-structured Pruning ($P_n$ in Table~\ref{tab:imagenet}).} We perform non-structured pruning on AlexNet and report the ratio of non-zero model parameters. 
\sys{} achieves higher accuracy with $3\%$ higher parameter size compared to a Bayesian Optimization approach~\cite{chen2018constraint}.

\noindent\textbf{Structured Pruning ($P_s$ in Table~\ref{tab:imagenet}).} On ResNet-50, \sys{} compresses the model to $1.5\times$ less FLOPs on average while achieving similar/higher test accuracy compared to best prior works. 
On VGG-16, \sys{} outperforms all heuristic methods and gives competing results with~\cite{liu2018rethinking} and the state-of-the-art RL method~\cite{he2018amc}. Note that~\cite{liu2018rethinking} does not propose a hyperparameter optimization algorithm and merely focuses on the training of already-compressed DNNs. As such, their approach is orthogonal to \sys{} and can be combined with our method to further improve final accuracy.

\noindent\textbf{Decomposition and Pruning ($P_s+D$ in Table~\ref{tab:imagenet}).} 
Using a combination of decomposition and pruning, \sys{} achieves $1.85\times$ lower FLOPs than related work (on average) with slightly higher accuracy on ResNet-50. On VGG-16, \sys{} pushes the state-of-the-art RL-based FLOPs reduction from $5\times$ to $7.1\times$ with higher accuracy.

\begin{table}[h]
    \centering
    \vspace{-0.2cm}
    \caption{Comparison with contemporary compression methods based on the non-zero parameter ratio/FLOPs.}\label{tab:cifar-10}\label{tab:imagenet}
    \vspace{-0.4cm}
    \includegraphics[width=0.95\columnwidth]{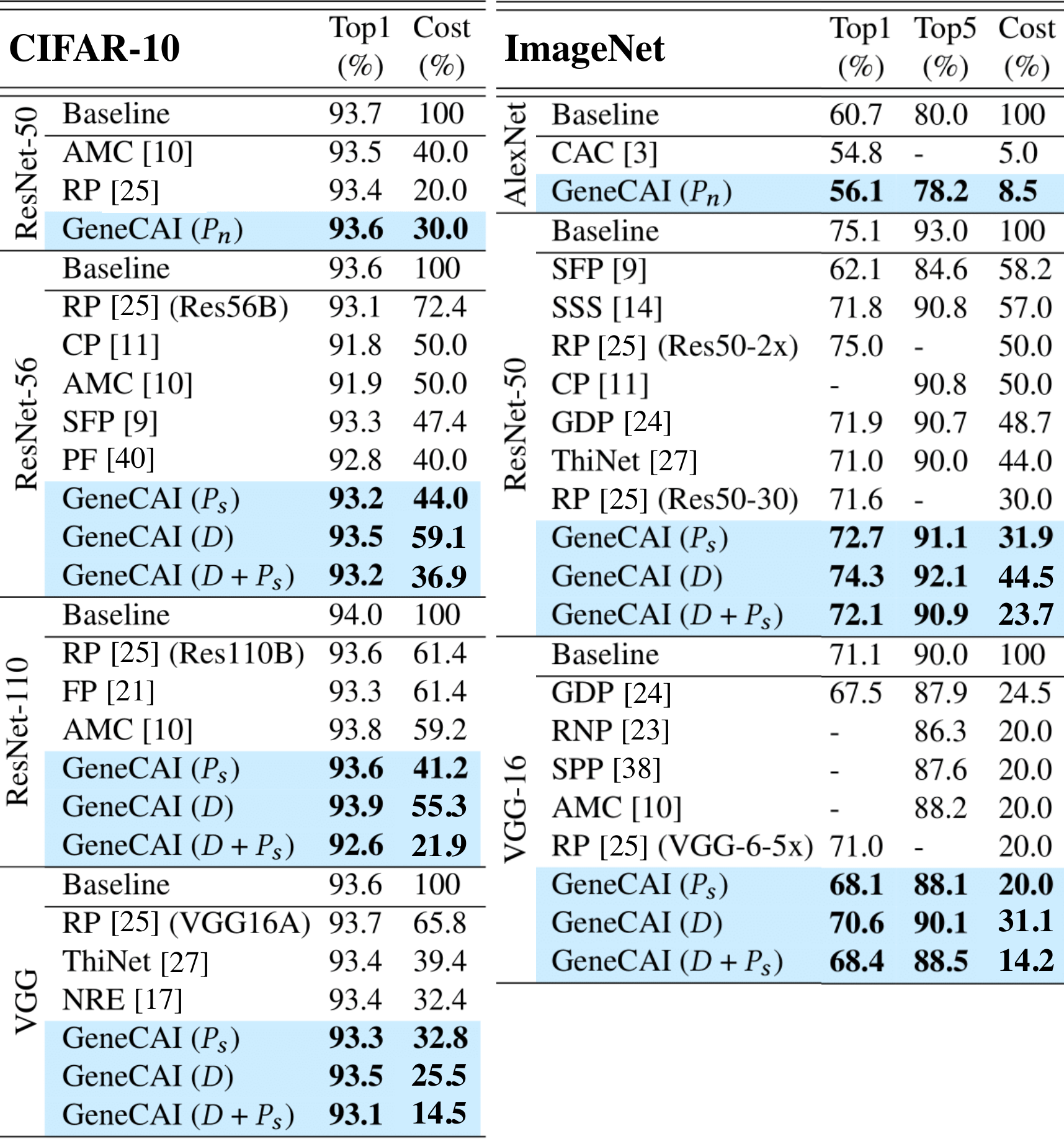}
    \vspace{-0.3cm}
\end{table}

%% file: 5_3_mobilenets.tex
\subsection{Compressing Compact Networks}\label{sec:mobilenet}
\vspace{-0.1cm}
To further demonstrate the effectiveness of \sys{} optimization, we apply compression to MobileNet architectures trained on ImageNet dataset. These networks are specifically designed for embedded applications with a strict efficiency constraint. As such, MobileNets inherently have very low complexity/redundancy which renders their compression quite challenging. We apply pruning to MobileNetV1 and MobileNetV2 with a population size of $50$, and let \sys{} run for $100$ iterations. We compare the compression rate and accuracy achieved by \sys{} with the FLOPs-accuracy Pareto curve of the original MobileNet architectures~\cite{howard2017mobilenets, sandler2018mobilenetv2}, the state-of-the-art AutoML approach~\cite{he2018amc,yang2018netadapt}, and compression-aware training methodologies, i.e., US-Nets~\cite{slimmable2019,Yu_2019_ICCV}.

\begin{table}[H]
\centering
\vspace{-0.2cm}
\caption{\sys{} pruning for MobileNetV1\&V2 on ImageNet.}\label{tab:mobilenet}
\vspace{-0.3cm}
\resizebox{0.82\columnwidth}{!}{
\begin{tabular}{clccc}
\hline
 & \multicolumn{1}{l}{Policy} & \begin{tabular}[c]{@{}c@{}}Top1 (\%)\end{tabular} & \begin{tabular}[c]{@{}c@{}}Top5 (\%)\end{tabular} & FLOPs \\ \hline \hline
& Baseline ($1\times$)              & 70.6                                                &   89.5    & 569 M                     \\ \cline{2-5}
& MobileNetV1 ($0.75\times$)~\cite{howard2017mobilenets}        & 68.4  & 88.2 & 325 M                     \\
& US-Nets~\cite{Yu_2019_ICCV}  & 69.5  & - & 325 M                     \\
\rowcolor{lightblue}
\multicolumn{1}{c}{\emptycell} & \sys{}  & \textbf{70.5} & 89.3  &   \textbf{323 M} \\ \cline{2-5}
& US-Nets~\cite{Yu_2019_ICCV}  & 68.8    & -  & 287 M       \\
& AMC~\cite{he2018amc}  & 70.5  & 89.1  & 285 M                     \\
& NetAdapt~\cite{yang2018netadapt} & 69.1  & - & 284 M                     \\
\rowcolor{lightblue}
\multicolumn{1}{c}{\emptycell} & \sys{}  & \textbf{70.4}  & 89.1 & \textbf{283 M}      \\ \cline{2-5}
& US-Nets~\cite{Yu_2019_ICCV}  & 66.8 & - & 217 M \\
\rowcolor{lightblue}
\multicolumn{1}{c}{\emptycell} & \sys{}  & \textbf{67.9}        & 88.1 & \textbf{210 M} \\ \cline{2-5}
& MobileNetV1 ($0.5\times$)~\cite{howard2017mobilenets}         & 63.7  & - & 149 M                     \\
& US-Nets~\cite{Yu_2019_ICCV} & 63.5  & - & 136 M                     \\
\rowcolor{lightblue}
\multirow{-13}{*}{\rotatebox[origin=c]{90}{\emptycell MobileNetV1}} & \sys{}  & \textbf{64.1}                         & 85.4 & \textbf{136 M}                    \\ \hline

& Baseline ($1\times$) & 71.6  & 90.3  & 313 M \\ \cline{2-5}
& MobileNetV2 ($0.75\times$)~\cite{sandler2018mobilenetv2}        & 69.8                                                      & 88.3                                                      & 220 M \\
& US-Nets~\cite{Yu_2019_ICCV} & 70.0                                                      & -                                                          & 222 M \\
& AMC~\cite{he2018amc}                        & -                                                         & 89.3                                                       & 220 M \\
\rowcolor{lightblue}
\multirow{-5}{*}{\rotatebox[origin=c]{90}{\emptycell MobileNetV2}} & \sys{}  & \textbf{70.1} & \textbf{89.5}           & \textbf{220 M} \\ \hline
\end{tabular}}
\end{table}

Table~\ref{tab:mobilenet} encloses the results of applying structured pruning to MobileNetV1 and MobileNetV2. We benchmark several target FLOPs and compare with prior work with similar computational complexities. On average, \sys{} achieves $1.2\%$ better accuracy than the MobileNetV1 Pareto curve. Compared to US-Nets, \sys{} achieves an average of $1.0\%$ higher accuracy. Under $\sim50\%$ FLOPs, \sys{} achieves $1.3\%$ higher accuracy than NetAdapt. Compared to AMC, \sys{} achieves lower FLOPs with comparable accuracy ($-0.1\%$). On MobileNetV2, for a $30\%$ FLOPs reduction, \sys{} achieves lower FLOPs and higher accuracy than US-Nets and higher accuracy with same FLOPs compared to AMC and MobileNetV2.

\noindent\textbf{Measured Speedup.} We present measured hardware speedups of \sys{} compressed MobileNets on an embedded GPU (NVIDIA Pascal) and an embedded CPU (ARM Cortex-A57) in Table~\ref{tab:mobilenet_speedup}. Measurements are averaged on $100$ runs using a batch size of $32$. \sys{} successfully models hardware cost to achieve real speedups on par with theory.

\begin{table}[h]
\centering
\vspace{-0.1cm}
\caption{\sys{} compressed MobileNets Speedup on embedded CPU and GPU for structured pruning on ImageNet.}\label{tab:mobilenet_speedup}
\vspace{-0.4cm}
\resizebox{0.95\columnwidth}{!}{
\begin{tabular}{lcbb}
\hline
\multicolumn{1}{c}{\multirow{2}{*}{Model}} & \multirow{2}{*}{\begin{tabular}[c]{@{}c@{}}Theoretical\\ Speedup\end{tabular}} & \multicolumn{2}{c}{Real Speedup}            \\ \cline{3-4} 
\multicolumn{1}{c}{} &                              & Cortex-A57 (CPU) & Pascal (GPU)  \\ \hline \hline
\multirow{4}{*}{MobileNetV1} 
& $1.7\times$   & $1.6\times$   & $1.3\times$ \\
& $2.0\times$  & $1.7\times$  & $1.4\times$ \\
& $2.7\times$  & $2.5\times$  & $1.7\times$ \\
& $4.0\times$  & $3.4\times$  & $1.9\times$  \\ \hline
MobileNetV2    & $1.4\times$                    & $1.4\times$   & $1.4\times$  \\ \hline
\end{tabular}}
\vspace{-0.5cm}
\end{table}

%% file: 6_conclusion.tex
\section{Conclusion}\label{sec:conclusion}
This paper introduces \sys{}, a method to automate DNN compression using genetic algorithms. Our algorithm learns how the compression hyperparameters should be set across layers to achieve a better performance than models designed by human experts. The core idea behind \sys{} is to translate compression hyperparameters into a vector of genes and explore the corresponding search-space using genetic operations.
This approach allows \sys{} to be generic and applicable to any combination of post-processing DNN compression methods. We showcase \sys{} effectiveness on four DNN compression methods, namely, structured and non-structured pruning, SVD, and Tucker decomposition. Experiments show that \sys{} achieves superior results compared to state-of-the-art heuristic and reinforcement-learning based algorithms on a variety of benchmarks. The proposed method is highly efficient as it does not require gradient computation and is entirely parallelizable.

\section{Acknowledgment}
This work was supported in part by Qualcomm Innovation Fellowship (QIF2019-US), NSF Grants CCF-1719133 and CCF-1513883, ARO  (W911NF1910317), and SRC-Auto (2019-AU-2899).

%% file: 0_main.bbl

\begin{thebibliography}{46}


\ifx \showCODEN    \undefined \def \showCODEN     #1{\unskip}     \fi
\ifx \showDOI      \undefined \def \showDOI       #1{#1}\fi
\ifx \showISBNx    \undefined \def \showISBNx     #1{\unskip}     \fi
\ifx \showISBNxiii \undefined \def \showISBNxiii  #1{\unskip}     \fi
\ifx \showISSN     \undefined \def \showISSN      #1{\unskip}     \fi
\ifx \showLCCN     \undefined \def \showLCCN      #1{\unskip}     \fi
\ifx \shownote     \undefined \def \shownote      #1{#1}          \fi
\ifx \showarticletitle \undefined \def \showarticletitle #1{#1}   \fi
\ifx \showURL      \undefined \def \showURL       {\relax}        \fi
\providecommand\bibfield[2]{#2}
\providecommand\bibinfo[2]{#2}
\providecommand\natexlab[1]{#1}
\providecommand\showeprint[2][]{arXiv:#2}

\bibitem[\protect\citeauthoryear{Alvarez and Salzmann}{Alvarez and
  Salzmann}{2017}]%
        {alvarez2017compression}
\bibfield{author}{\bibinfo{person}{Jose~M Alvarez} {and}
  \bibinfo{person}{Mathieu Salzmann}.} \bibinfo{year}{2017}\natexlab{}.
\newblock \showarticletitle{Compression-aware training of deep networks}. In
  \bibinfo{booktitle}{{\em NIPS}}. \bibinfo{pages}{856--867}.
\newblock


\bibitem[\protect\citeauthoryear{B{\"a}ck, Fogel, and Michalewicz}{B{\"a}ck
  et~al\mbox{.}}{1997}]%
        {back1997handbook}
\bibfield{author}{\bibinfo{person}{Thomas B{\"a}ck}, \bibinfo{person}{David~B
  Fogel}, {and} \bibinfo{person}{Zbigniew Michalewicz}.}
  \bibinfo{year}{1997}\natexlab{}.
\newblock \bibinfo{booktitle}{{\em Handbook of evolutionary computation}}.
\newblock \bibinfo{publisher}{CRC Press}.
\newblock


\bibitem[\protect\citeauthoryear{Chen et~al\mbox{.}}{Chen
  et~al\mbox{.}}{2018}]%
        {chen2018constraint}
\bibfield{author}{\bibinfo{person}{Changan Chen} {et~al\mbox{.}}}
  \bibinfo{year}{2018}\natexlab{}.
\newblock \showarticletitle{Constraint-aware deep neural network compression}.
  In \bibinfo{booktitle}{{\em ECCV}}. \bibinfo{pages}{400--415}.
\newblock


\bibitem[\protect\citeauthoryear{Chong and Zak}{Chong and Zak}{2013}]%
        {chong2013introduction}
\bibfield{author}{\bibinfo{person}{Edwin~KP Chong} {and}
  \bibinfo{person}{Stanislaw~H Zak}.} \bibinfo{year}{2013}\natexlab{}.
\newblock \bibinfo{booktitle}{{\em An introduction to optimization}}.
  Vol.~\bibinfo{volume}{76}.
\newblock \bibinfo{publisher}{John Wiley \& Sons}.
\newblock


\bibitem[\protect\citeauthoryear{Ghasemzadeh, Samragh, and
  Koushanfar}{Ghasemzadeh et~al\mbox{.}}{2018}]%
        {ghasemzadeh2018rebnet}
\bibfield{author}{\bibinfo{person}{Mohammad Ghasemzadeh},
  \bibinfo{person}{Mohammad Samragh}, {and} \bibinfo{person}{Farinaz
  Koushanfar}.} \bibinfo{year}{2018}\natexlab{}.
\newblock \showarticletitle{Rebnet: Residual binarized neural network}. In
  \bibinfo{booktitle}{{\em 2018 IEEE 26th Annual International Symposium on
  Field-Programmable Custom Computing Machines (FCCM)}}. IEEE,
  \bibinfo{pages}{57--64}.
\newblock


\bibitem[\protect\citeauthoryear{Han et~al\mbox{.}}{Han et~al\mbox{.}}{2015a}]%
        {han2015learning}
\bibfield{author}{\bibinfo{person}{Song Han} {et~al\mbox{.}}}
  \bibinfo{year}{2015}\natexlab{a}.
\newblock \showarticletitle{Learning both weights and connections for efficient
  neural network}. In \bibinfo{booktitle}{{\em NIPS}}.
  \bibinfo{pages}{1135--1143}.
\newblock


\bibitem[\protect\citeauthoryear{Han, Mao, and Dally}{Han
  et~al\mbox{.}}{2015b}]%
        {han2015deep}
\bibfield{author}{\bibinfo{person}{Song Han}, \bibinfo{person}{Huizi Mao},
  {and} \bibinfo{person}{William~J Dally}.} \bibinfo{year}{2015}\natexlab{b}.
\newblock \showarticletitle{Deep compression: Compressing deep neural networks
  with pruning, trained quantization and huffman coding}.
\newblock \bibinfo{journal}{{\em arXiv preprint arXiv:1510.00149\/}}
  (\bibinfo{year}{2015}).
\newblock


\bibitem[\protect\citeauthoryear{He et~al\mbox{.}}{He et~al\mbox{.}}{2016}]%
        {he2016deep}
\bibfield{author}{\bibinfo{person}{Kaiming He} {et~al\mbox{.}}}
  \bibinfo{year}{2016}\natexlab{}.
\newblock \showarticletitle{Deep residual learning for image recognition}. In
  \bibinfo{booktitle}{{\em Proceedings of the IEEE conference on computer
  vision and pattern recognition}}. \bibinfo{pages}{770--778}.
\newblock


\bibitem[\protect\citeauthoryear{He, Kang, Dong, Fu, and Yang}{He
  et~al\mbox{.}}{2018a}]%
        {he2018soft}
\bibfield{author}{\bibinfo{person}{Yang He}, \bibinfo{person}{Guoliang Kang},
  \bibinfo{person}{Xuanyi Dong}, \bibinfo{person}{Yanwei Fu}, {and}
  \bibinfo{person}{Yi Yang}.} \bibinfo{year}{2018}\natexlab{a}.
\newblock \showarticletitle{Soft filter pruning for accelerating deep
  convolutional neural networks}.
\newblock \bibinfo{journal}{{\em arXiv preprint arXiv:1808.06866\/}}
  (\bibinfo{year}{2018}).
\newblock


\bibitem[\protect\citeauthoryear{He, Lin, Liu, Wang, Li, and Han}{He
  et~al\mbox{.}}{2018b}]%
        {he2018amc}
\bibfield{author}{\bibinfo{person}{Yihui He}, \bibinfo{person}{Ji Lin},
  \bibinfo{person}{Zhijian Liu}, \bibinfo{person}{Hanrui Wang},
  \bibinfo{person}{Li-Jia Li}, {and} \bibinfo{person}{Song Han}.}
  \bibinfo{year}{2018}\natexlab{b}.
\newblock \showarticletitle{Amc: Automl for model compression and acceleration
  on mobile devices}. In \bibinfo{booktitle}{{\em Proceedings of the European
  Conference on Computer Vision (ECCV)}}. \bibinfo{pages}{784--800}.
\newblock


\bibitem[\protect\citeauthoryear{He, Zhang, and Sun}{He et~al\mbox{.}}{2017}]%
        {he2017channel}
\bibfield{author}{\bibinfo{person}{Yihui He}, \bibinfo{person}{Xiangyu Zhang},
  {and} \bibinfo{person}{Jian Sun}.} \bibinfo{year}{2017}\natexlab{}.
\newblock \showarticletitle{Channel pruning for accelerating very deep neural
  networks}. In \bibinfo{booktitle}{{\em International Conference on Computer
  Vision (ICCV)}}, Vol.~\bibinfo{volume}{2}.
\newblock


\bibitem[\protect\citeauthoryear{Howard, Zhu, Chen, Kalenichenko, Wang, Weyand,
  Andreetto, and Adam}{Howard et~al\mbox{.}}{2017}]%
        {howard2017mobilenets}
\bibfield{author}{\bibinfo{person}{Andrew~G Howard}, \bibinfo{person}{Menglong
  Zhu}, \bibinfo{person}{Bo Chen}, \bibinfo{person}{Dmitry Kalenichenko},
  \bibinfo{person}{Weijun Wang}, \bibinfo{person}{Tobias Weyand},
  \bibinfo{person}{Marco Andreetto}, {and} \bibinfo{person}{Hartwig Adam}.}
  \bibinfo{year}{2017}\natexlab{}.
\newblock \showarticletitle{Mobilenets: Efficient convolutional neural networks
  for mobile vision applications}.
\newblock \bibinfo{journal}{{\em arXiv preprint arXiv:1704.04861\/}}
  (\bibinfo{year}{2017}).
\newblock


\bibitem[\protect\citeauthoryear{Hu, Sun, Li, Wang, and Gu}{Hu
  et~al\mbox{.}}{2018}]%
        {hu2018novel}
\bibfield{author}{\bibinfo{person}{Yiming Hu}, \bibinfo{person}{Siyang Sun},
  \bibinfo{person}{Jianquan Li}, \bibinfo{person}{Xingang Wang}, {and}
  \bibinfo{person}{Qingyi Gu}.} \bibinfo{year}{2018}\natexlab{}.
\newblock \showarticletitle{A novel channel pruning method for deep neural
  network compression}.
\newblock \bibinfo{journal}{{\em arXiv preprint arXiv:1805.11394\/}}
  (\bibinfo{year}{2018}).
\newblock


\bibitem[\protect\citeauthoryear{Huang and Wang}{Huang and Wang}{2018}]%
        {huang2018data}
\bibfield{author}{\bibinfo{person}{Zehao Huang} {and} \bibinfo{person}{Naiyan
  Wang}.} \bibinfo{year}{2018}\natexlab{}.
\newblock \showarticletitle{Data-driven sparse structure selection for deep
  neural networks}. In \bibinfo{booktitle}{{\em Proceedings of the European
  Conference on Computer Vision (ECCV)}}. \bibinfo{pages}{304--320}.
\newblock


\bibitem[\protect\citeauthoryear{Javaheripi, Rouhani, and
  Koushanfar}{Javaheripi et~al\mbox{.}}{2019a}]%
        {javaheripi2019swnet}
\bibfield{author}{\bibinfo{person}{Mojan Javaheripi},
  \bibinfo{person}{Bita~Darvish Rouhani}, {and} \bibinfo{person}{Farinaz
  Koushanfar}.} \bibinfo{year}{2019}\natexlab{a}.
\newblock \showarticletitle{SWNet: Small-World Neural Networks and Rapid
  Convergence}.
\newblock \bibinfo{journal}{{\em arXiv preprint arXiv:1904.04862\/}}
  (\bibinfo{year}{2019}).
\newblock


\bibitem[\protect\citeauthoryear{Javaheripi, Samragh, and
  Koushanfar}{Javaheripi et~al\mbox{.}}{2019b}]%
        {javaheripi2019peeking}
\bibfield{author}{\bibinfo{person}{Mojan Javaheripi}, \bibinfo{person}{Mohammad
  Samragh}, {and} \bibinfo{person}{Farinaz Koushanfar}.}
  \bibinfo{year}{2019}\natexlab{b}.
\newblock \showarticletitle{Peeking Into the Black Box: A Tutorial on Automated
  Design Optimization and Parameter Search}.
\newblock \bibinfo{journal}{{\em IEEE Solid-State Circuits Magazine\/}}
  \bibinfo{volume}{11}, \bibinfo{number}{4} (\bibinfo{year}{2019}),
  \bibinfo{pages}{23--28}.
\newblock


\bibitem[\protect\citeauthoryear{Jiang, Li, Qian, and Tang}{Jiang
  et~al\mbox{.}}{2018}]%
        {jiang2018efficient}
\bibfield{author}{\bibinfo{person}{Chunhui Jiang}, \bibinfo{person}{Guiying
  Li}, \bibinfo{person}{Chao Qian}, {and} \bibinfo{person}{Ke Tang}.}
  \bibinfo{year}{2018}\natexlab{}.
\newblock \showarticletitle{Efficient DNN Neuron Pruning by Minimizing
  Layer-wise Nonlinear Reconstruction Error}. In \bibinfo{booktitle}{{\em
  IJCAI}}. \bibinfo{pages}{2--2}.
\newblock


\bibitem[\protect\citeauthoryear{Kim, Park, Yoo, Choi, Yang, and Shin}{Kim
  et~al\mbox{.}}{2015}]%
        {kim2015compression}
\bibfield{author}{\bibinfo{person}{Yong-Deok Kim}, \bibinfo{person}{Eunhyeok
  Park}, \bibinfo{person}{Sungjoo Yoo}, \bibinfo{person}{Taelim Choi},
  \bibinfo{person}{Lu Yang}, {and} \bibinfo{person}{Dongjun Shin}.}
  \bibinfo{year}{2015}\natexlab{}.
\newblock \showarticletitle{Compression of deep convolutional neural networks
  for fast and low power mobile applications}.
\newblock \bibinfo{journal}{{\em arXiv preprint arXiv:1511.06530\/}}
  (\bibinfo{year}{2015}).
\newblock


\bibitem[\protect\citeauthoryear{Krizhevsky et~al\mbox{.}}{Krizhevsky
  et~al\mbox{.}}{2012}]%
        {krizhevsky2012imagenet}
\bibfield{author}{\bibinfo{person}{Alex Krizhevsky} {et~al\mbox{.}}}
  \bibinfo{year}{2012}\natexlab{}.
\newblock \showarticletitle{Imagenet classification with deep convolutional
  neural networks}. In \bibinfo{booktitle}{{\em Advances in neural information
  processing systems}}. \bibinfo{pages}{1097--1105}.
\newblock


\bibitem[\protect\citeauthoryear{Krizhevsky and Hinton}{Krizhevsky and
  Hinton}{2009}]%
        {krizhevsky2009learning}
\bibfield{author}{\bibinfo{person}{Alex Krizhevsky} {and}
  \bibinfo{person}{Geoffrey Hinton}.} \bibinfo{year}{2009}\natexlab{}.
\newblock \bibinfo{booktitle}{{\em Learning multiple layers of features from
  tiny images}}.
\newblock \bibinfo{type}{{T}echnical {R}eport}.
  \bibinfo{institution}{Citeseer}.
\newblock


\bibitem[\protect\citeauthoryear{Li, Kadav, Durdanovic, Samet, and Graf}{Li
  et~al\mbox{.}}{2016}]%
        {li2016pruning}
\bibfield{author}{\bibinfo{person}{Hao Li}, \bibinfo{person}{Asim Kadav},
  \bibinfo{person}{Igor Durdanovic}, \bibinfo{person}{Hanan Samet}, {and}
  \bibinfo{person}{Hans~Peter Graf}.} \bibinfo{year}{2016}\natexlab{}.
\newblock \showarticletitle{Pruning filters for efficient convnets}.
\newblock \bibinfo{journal}{{\em arXiv preprint arXiv:1608.08710\/}}
  (\bibinfo{year}{2016}).
\newblock


\bibitem[\protect\citeauthoryear{Liang, Meyerson, Hodjat, Fink, Mutch, and
  Miikkulainen}{Liang et~al\mbox{.}}{2019}]%
        {liang2019evolutionary}
\bibfield{author}{\bibinfo{person}{Jason Liang}, \bibinfo{person}{Elliot
  Meyerson}, \bibinfo{person}{Babak Hodjat}, \bibinfo{person}{Dan Fink},
  \bibinfo{person}{Karl Mutch}, {and} \bibinfo{person}{Risto Miikkulainen}.}
  \bibinfo{year}{2019}\natexlab{}.
\newblock \showarticletitle{Evolutionary neural automl for deep learning}. In
  \bibinfo{booktitle}{{\em Proceedings of the Genetic and Evolutionary
  Computation Conference}}. \bibinfo{pages}{401--409}.
\newblock


\bibitem[\protect\citeauthoryear{Lin, Rao, Lu, and Zhou}{Lin
  et~al\mbox{.}}{2017}]%
        {lin2017runtime}
\bibfield{author}{\bibinfo{person}{Ji Lin}, \bibinfo{person}{Yongming Rao},
  \bibinfo{person}{Jiwen Lu}, {and} \bibinfo{person}{Jie Zhou}.}
  \bibinfo{year}{2017}\natexlab{}.
\newblock \showarticletitle{Runtime neural pruning}. In
  \bibinfo{booktitle}{{\em Advances in Neural Information Processing Systems}}.
  \bibinfo{pages}{2181--2191}.
\newblock


\bibitem[\protect\citeauthoryear{Lin, Ji, Li, Wu, Huang, and Zhang}{Lin
  et~al\mbox{.}}{2018}]%
        {lin2018accelerating}
\bibfield{author}{\bibinfo{person}{Shaohui Lin}, \bibinfo{person}{Rongrong Ji},
  \bibinfo{person}{Yuchao Li}, \bibinfo{person}{Yongjian Wu},
  \bibinfo{person}{Feiyue Huang}, {and} \bibinfo{person}{Baochang Zhang}.}
  \bibinfo{year}{2018}\natexlab{}.
\newblock \showarticletitle{Accelerating Convolutional Networks via Global \&
  Dynamic Filter Pruning.}. In \bibinfo{booktitle}{{\em IJCAI}}.
  \bibinfo{pages}{2425--2432}.
\newblock


\bibitem[\protect\citeauthoryear{Liu et~al\mbox{.}}{Liu et~al\mbox{.}}{2018}]%
        {liu2018rethinking}
\bibfield{author}{\bibinfo{person}{Zhuang Liu} {et~al\mbox{.}}}
  \bibinfo{year}{2018}\natexlab{}.
\newblock \showarticletitle{Rethinking the value of network pruning}.
\newblock \bibinfo{journal}{{\em arXiv:1810.05270\/}} (\bibinfo{year}{2018}).
\newblock


\bibitem[\protect\citeauthoryear{Lu, Whalen, Boddeti, Dhebar, Deb, Goodman, and
  Banzhaf}{Lu et~al\mbox{.}}{2019}]%
        {lu2019nsga}
\bibfield{author}{\bibinfo{person}{Zhichao Lu}, \bibinfo{person}{Ian Whalen},
  \bibinfo{person}{Vishnu Boddeti}, \bibinfo{person}{Yashesh Dhebar},
  \bibinfo{person}{Kalyanmoy Deb}, \bibinfo{person}{Erik Goodman}, {and}
  \bibinfo{person}{Wolfgang Banzhaf}.} \bibinfo{year}{2019}\natexlab{}.
\newblock \showarticletitle{NSGA-Net: neural architecture search using
  multi-objective genetic algorithm}. In \bibinfo{booktitle}{{\em Proceedings
  of the Genetic and Evolutionary Computation Conference}}.
  \bibinfo{pages}{419--427}.
\newblock


\bibitem[\protect\citeauthoryear{Luo, Wu, and Lin}{Luo et~al\mbox{.}}{2017}]%
        {luo2017thinet}
\bibfield{author}{\bibinfo{person}{Jian-Hao Luo}, \bibinfo{person}{Jianxin Wu},
  {and} \bibinfo{person}{Weiyao Lin}.} \bibinfo{year}{2017}\natexlab{}.
\newblock \showarticletitle{Thinet: A filter level pruning method for deep
  neural network compression}. In \bibinfo{booktitle}{{\em Proceedings of the
  IEEE international conference on computer vision}}.
  \bibinfo{pages}{5058--5066}.
\newblock


\bibitem[\protect\citeauthoryear{Molchanov, Tyree, Karras, Aila, and
  Kautz}{Molchanov et~al\mbox{.}}{2016}]%
        {molchanov2016pruning}
\bibfield{author}{\bibinfo{person}{Pavlo Molchanov}, \bibinfo{person}{Stephen
  Tyree}, \bibinfo{person}{Tero Karras}, \bibinfo{person}{Timo Aila}, {and}
  \bibinfo{person}{Jan Kautz}.} \bibinfo{year}{2016}\natexlab{}.
\newblock \showarticletitle{Pruning convolutional neural networks for resource
  efficient transfer learning}.
\newblock \bibinfo{journal}{{\em arXiv preprint arXiv:1611.06440\/}}
  \bibinfo{volume}{3} (\bibinfo{year}{2016}).
\newblock


\bibitem[\protect\citeauthoryear{Real, Moore, Selle, Saxena, Suematsu, Tan, Le,
  and Kurakin}{Real et~al\mbox{.}}{2017}]%
        {real2017large}
\bibfield{author}{\bibinfo{person}{Esteban Real}, \bibinfo{person}{Sherry
  Moore}, \bibinfo{person}{Andrew Selle}, \bibinfo{person}{Saurabh Saxena},
  \bibinfo{person}{Yutaka~Leon Suematsu}, \bibinfo{person}{Jie Tan},
  \bibinfo{person}{Quoc~V Le}, {and} \bibinfo{person}{Alexey Kurakin}.}
  \bibinfo{year}{2017}\natexlab{}.
\newblock \showarticletitle{Large-scale evolution of image classifiers}. In
  \bibinfo{booktitle}{{\em Proceedings of the 34th International Conference on
  Machine Learning-Volume 70}}. JMLR. org, \bibinfo{pages}{2902--2911}.
\newblock


\bibitem[\protect\citeauthoryear{Russakovsky, Deng, Su, Krause, Satheesh, Ma,
  Huang, Karpathy, Khosla, Bernstein, et~al\mbox{.}}{Russakovsky
  et~al\mbox{.}}{2015}]%
        {russakovsky2015imagenet}
\bibfield{author}{\bibinfo{person}{Olga Russakovsky}, \bibinfo{person}{Jia
  Deng}, \bibinfo{person}{Hao Su}, \bibinfo{person}{Jonathan Krause},
  \bibinfo{person}{Sanjeev Satheesh}, \bibinfo{person}{Sean Ma},
  \bibinfo{person}{Zhiheng Huang}, \bibinfo{person}{Andrej Karpathy},
  \bibinfo{person}{Aditya Khosla}, \bibinfo{person}{Michael Bernstein},
  {et~al\mbox{.}}} \bibinfo{year}{2015}\natexlab{}.
\newblock \showarticletitle{Imagenet large scale visual recognition challenge}.
\newblock \bibinfo{journal}{{\em International journal of computer vision\/}}
  \bibinfo{volume}{115}, \bibinfo{number}{3} (\bibinfo{year}{2015}),
  \bibinfo{pages}{211--252}.
\newblock


\bibitem[\protect\citeauthoryear{Salimans, Ho, Chen, Sidor, and
  Sutskever}{Salimans et~al\mbox{.}}{2017}]%
        {salimans2017evolution}
\bibfield{author}{\bibinfo{person}{Tim Salimans}, \bibinfo{person}{Jonathan
  Ho}, \bibinfo{person}{Xi Chen}, \bibinfo{person}{Szymon Sidor}, {and}
  \bibinfo{person}{Ilya Sutskever}.} \bibinfo{year}{2017}\natexlab{}.
\newblock \showarticletitle{Evolution strategies as a scalable alternative to
  reinforcement learning}.
\newblock \bibinfo{journal}{{\em arXiv preprint arXiv:1703.03864\/}}
  (\bibinfo{year}{2017}).
\newblock


\bibitem[\protect\citeauthoryear{Samragh, Ghasemzadeh, and Koushanfar}{Samragh
  et~al\mbox{.}}{2017}]%
        {samragh2017customizing}
\bibfield{author}{\bibinfo{person}{Mohammad Samragh}, \bibinfo{person}{Mohammad
  Ghasemzadeh}, {and} \bibinfo{person}{Farinaz Koushanfar}.}
  \bibinfo{year}{2017}\natexlab{}.
\newblock \showarticletitle{Customizing neural networks for efficient fpga
  implementation}. In \bibinfo{booktitle}{{\em Field-Programmable Custom
  Computing Machines (FCCM), 2017 IEEE 25th Annual International Symposium
  on}}. IEEE, \bibinfo{pages}{85--92}.
\newblock


\bibitem[\protect\citeauthoryear{Samragh, Javaheripi, and Koushanfar}{Samragh
  et~al\mbox{.}}{2019a}]%
        {samragh2019autorank}
\bibfield{author}{\bibinfo{person}{Mohammad Samragh}, \bibinfo{person}{Mojan
  Javaheripi}, {and} \bibinfo{person}{Farinaz Koushanfar}.}
  \bibinfo{year}{2019}\natexlab{a}.
\newblock \showarticletitle{AutoRank: Automated rank selection for effective
  neural network customization}. In \bibinfo{booktitle}{{\em Proceedings of the
  ML-for-Systems Workshop at the 46th International Symposium on Computer
  Architecture (ISCA’19)}}.
\newblock


\bibitem[\protect\citeauthoryear{Samragh, Javaheripi, and Koushanfar}{Samragh
  et~al\mbox{.}}{2019b}]%
        {samragh2019codex}
\bibfield{author}{\bibinfo{person}{Mohammad Samragh}, \bibinfo{person}{Mojan
  Javaheripi}, {and} \bibinfo{person}{Farinaz Koushanfar}.}
  \bibinfo{year}{2019}\natexlab{b}.
\newblock \showarticletitle{CodeX: Bit-Flexible Encoding for Streaming-based
  FPGA Acceleration of DNNs}.
\newblock \bibinfo{journal}{{\em arXiv preprint arXiv:1901.05582\/}}
  (\bibinfo{year}{2019}).
\newblock


\bibitem[\protect\citeauthoryear{Sandler, Howard, Zhu, Zhmoginov, and
  Chen}{Sandler et~al\mbox{.}}{2018}]%
        {sandler2018mobilenetv2}
\bibfield{author}{\bibinfo{person}{Mark Sandler}, \bibinfo{person}{Andrew
  Howard}, \bibinfo{person}{Menglong Zhu}, \bibinfo{person}{Andrey Zhmoginov},
  {and} \bibinfo{person}{Liang-Chieh Chen}.} \bibinfo{year}{2018}\natexlab{}.
\newblock \showarticletitle{Mobilenetv2: Inverted residuals and linear
  bottlenecks}. In \bibinfo{booktitle}{{\em Proceedings of the IEEE Conference
  on Computer Vision and Pattern Recognition}}. \bibinfo{pages}{4510--4520}.
\newblock


\bibitem[\protect\citeauthoryear{Simonyan and Zisserman}{Simonyan and
  Zisserman}{2014}]%
        {simonyan2014very}
\bibfield{author}{\bibinfo{person}{Karen Simonyan} {and}
  \bibinfo{person}{Andrew Zisserman}.} \bibinfo{year}{2014}\natexlab{}.
\newblock \showarticletitle{Very deep convolutional networks for large-scale
  image recognition}.
\newblock \bibinfo{journal}{{\em arXiv preprint arXiv:1409.1556\/}}
  (\bibinfo{year}{2014}).
\newblock


\bibitem[\protect\citeauthoryear{Wang, Sun, Xue, and Zhang}{Wang
  et~al\mbox{.}}{2019}]%
        {wang2019evolving}
\bibfield{author}{\bibinfo{person}{Bin Wang}, \bibinfo{person}{Yanan Sun},
  \bibinfo{person}{Bing Xue}, {and} \bibinfo{person}{Mengjie Zhang}.}
  \bibinfo{year}{2019}\natexlab{}.
\newblock \showarticletitle{Evolving deep neural networks by multi-objective
  particle swarm optimization for image classification}. In
  \bibinfo{booktitle}{{\em Proceedings of the Genetic and Evolutionary
  Computation Conference}}. \bibinfo{pages}{490--498}.
\newblock


\bibitem[\protect\citeauthoryear{Wang, Zhang, Wang, and Hu}{Wang
  et~al\mbox{.}}{2017}]%
        {wang2017structured}
\bibfield{author}{\bibinfo{person}{Huan Wang}, \bibinfo{person}{Qiming Zhang},
  \bibinfo{person}{Yuehai Wang}, {and} \bibinfo{person}{Haoji Hu}.}
  \bibinfo{year}{2017}\natexlab{}.
\newblock \showarticletitle{Structured probabilistic pruning for convolutional
  neural network acceleration}.
\newblock \bibinfo{journal}{{\em arXiv preprint arXiv:1709.06994\/}}
  (\bibinfo{year}{2017}).
\newblock


\bibitem[\protect\citeauthoryear{Wang et~al\mbox{.}}{Wang
  et~al\mbox{.}}{2018}]%
        {wang2018haq}
\bibfield{author}{\bibinfo{person}{Kuan Wang} {et~al\mbox{.}}}
  \bibinfo{year}{2018}\natexlab{}.
\newblock \showarticletitle{HAQ: hardware-aware automated quantization}.
\newblock \bibinfo{journal}{{\em arXiv preprint arXiv:1811.08886\/}}
  (\bibinfo{year}{2018}).
\newblock


\bibitem[\protect\citeauthoryear{Xie and Yuille}{Xie and Yuille}{2017}]%
        {xie2017genetic}
\bibfield{author}{\bibinfo{person}{Lingxi Xie} {and} \bibinfo{person}{Alan
  Yuille}.} \bibinfo{year}{2017}\natexlab{}.
\newblock \showarticletitle{Genetic cnn}.
\newblock \bibinfo{journal}{{\em arXiv preprint arXiv:1703.01513\/}}
  (\bibinfo{year}{2017}).
\newblock


\bibitem[\protect\citeauthoryear{Yang, Howard, Chen, Zhang, Go, Sandler, Sze,
  and Adam}{Yang et~al\mbox{.}}{2018}]%
        {yang2018netadapt}
\bibfield{author}{\bibinfo{person}{Tien-Ju Yang}, \bibinfo{person}{Andrew
  Howard}, \bibinfo{person}{Bo Chen}, \bibinfo{person}{Xiao Zhang},
  \bibinfo{person}{Alec Go}, \bibinfo{person}{Mark Sandler},
  \bibinfo{person}{Vivienne Sze}, {and} \bibinfo{person}{Hartwig Adam}.}
  \bibinfo{year}{2018}\natexlab{}.
\newblock \showarticletitle{Netadapt: Platform-aware neural network adaptation
  for mobile applications}. In \bibinfo{booktitle}{{\em Proceedings of the
  European Conference on Computer Vision (ECCV)}}. \bibinfo{pages}{285--300}.
\newblock


\bibitem[\protect\citeauthoryear{Yazdanbakhsh et~al\mbox{.}}{Yazdanbakhsh
  et~al\mbox{.}}{2018}]%
        {yazdanbakhsh2018releq}
\bibfield{author}{\bibinfo{person}{Amir Yazdanbakhsh} {et~al\mbox{.}}}
  \bibinfo{year}{2018}\natexlab{}.
\newblock \showarticletitle{ReLeQ: An Automatic Reinforcement Learning Approach
  for Deep Quantization of Neural Networks}.
\newblock \bibinfo{journal}{{\em arXiv preprint arXiv:1811.01704\/}}
  (\bibinfo{year}{2018}).
\newblock


\bibitem[\protect\citeauthoryear{Yu and Huang}{Yu and Huang}{2019}]%
        {Yu_2019_ICCV}
\bibfield{author}{\bibinfo{person}{Jiahui Yu} {and} \bibinfo{person}{Thomas~S.
  Huang}.} \bibinfo{year}{2019}\natexlab{}.
\newblock \showarticletitle{Universally Slimmable Networks and Improved
  Training Techniques}. In \bibinfo{booktitle}{{\em The IEEE International
  Conference on Computer Vision (ICCV)}}.
\newblock


\bibitem[\protect\citeauthoryear{Yu, Yang, Xu, Yang, and Huang}{Yu
  et~al\mbox{.}}{2019}]%
        {slimmable2019}
\bibfield{author}{\bibinfo{person}{Jiahui Yu}, \bibinfo{person}{Linjie Yang},
  \bibinfo{person}{Ning Xu}, \bibinfo{person}{Jianchao Yang}, {and}
  \bibinfo{person}{{Thomas S} Huang}.} \bibinfo{year}{2019}\natexlab{}.
\newblock \showarticletitle{Slimmable neural networks}.
\newblock
\newblock
\shownote{7th International Conference on Learning Representations, ICLR 2019 ;
  Conference date: 06-05-2019 Through 09-05-2019.}


\bibitem[\protect\citeauthoryear{Zhou, Alvarez, and Porikli}{Zhou
  et~al\mbox{.}}{2016a}]%
        {zhou2016less}
\bibfield{author}{\bibinfo{person}{Hao Zhou}, \bibinfo{person}{Jose~M Alvarez},
  {and} \bibinfo{person}{Fatih Porikli}.} \bibinfo{year}{2016}\natexlab{a}.
\newblock \showarticletitle{Less is more: Towards compact cnns}. In
  \bibinfo{booktitle}{{\em European Conference on Computer Vision}}. Springer,
  \bibinfo{pages}{662--677}.
\newblock


\bibitem[\protect\citeauthoryear{Zhou, Wu, Ni, Zhou, Wen, and Zou}{Zhou
  et~al\mbox{.}}{2016b}]%
        {zhou2016dorefa}
\bibfield{author}{\bibinfo{person}{Shuchang Zhou}, \bibinfo{person}{Yuxin Wu},
  \bibinfo{person}{Zekun Ni}, \bibinfo{person}{Xinyu Zhou}, \bibinfo{person}{He
  Wen}, {and} \bibinfo{person}{Yuheng Zou}.} \bibinfo{year}{2016}\natexlab{b}.
\newblock \showarticletitle{Dorefa-net: Training low bitwidth convolutional
  neural networks with low bitwidth gradients}.
\newblock \bibinfo{journal}{{\em arXiv preprint arXiv:1606.06160\/}}
  (\bibinfo{year}{2016}).
\newblock


\end{thebibliography}
